\PassOptionsToPackage{dvipsnames}{xcolor}
\documentclass[pmlr,twocolumn,10pt]{jmlr} 

\usepackage{booktabs}
\usepackage{siunitx}

\usepackage[switch]{lineno}
\usepackage{graphicx}
\usepackage{xcolor} 
\usepackage{makecell}
\usepackage{stfloats}
\usepackage{float}
\usepackage{listings}
\usepackage{soul}          
\usepackage[normalem]{ulem}

\usepackage{booktabs}

\definecolor{cbblue}{HTML}{0173B2}
\definecolor{cborange}{HTML}{DE8F05}

\newcommand{\xhdr}[1]{\vspace{1.6mm}\noindent{{\bf #1.}}}

\theorembodyfont{\upshape}
\theoremheaderfont{\scshape}
\theorempostheader{:}
\theoremsep{\newline}

\jmlrvolume{333}
\jmlryear{2026}
\jmlrsubmitted{LEAVE UNSET}
\jmlrpublished{LEAVE UNSET}
\jmlrworkshop{Conference on Health, Inference, and Learning (CHIL) 2026} 

\title[Training-Free Adaptation of New-Generation LLMs using Legacy Clinical Models]{Training-Free Adaptation of New-Generation LLMs \titlebreak using Legacy Clinical Models}

\author{%
\Name{Sasha Ronaghi}\Email{sronaghi@stanford.edu}\\
\addr Stanford University
\AND
\Name{Chloe Stanwyck}\Email{chloeo@stanford.edu}\\
\addr Stanford University
\AND
\Name{Asad Aali}\Email{asadaali@stanford.edu}\\
\addr Stanford University
\AND
\Name{Amir Ronaghi}\Email{aronaghi1@gmail.com}\\
\addr MemorialCare
\AND
\Name{Miguel Angel Fuentes Hernandez}\Email{migufuen@stanford.edu}\\
\addr Stanford University
\AND
\Name{Tina Hernandez Boussard}\Email{boussard@stanford.edu}\\
\addr Stanford University
\AND
\Name{Emily Alsentzer}\Email{ealsentzer@stanford.edu}\\
\addr Stanford University
}

\begin{document}

\maketitle

\begin{abstract}
Adapting language models to the clinical domain through continued pretraining and instruction tuning requires costly retraining for each new model generation. We propose \textit{Cross-Architecture Proxy Tuning} (CAPT), a model-ensembling approach
that enables training-free adaptation of state-of-the-art general-domain models using existing clinical models. CAPT supports models with disjoint vocabularies, leveraging contrastive decoding to selectively inject clinically relevant signals while preserving the general-domain model's reasoning and fluency. On six clinical classification and text-generation tasks, CAPT with a new-generation general-domain model and an older-generation clinical model consistently outperforms both models individually and state-of-the-art ensembling approaches (average +17.6\% over UniTE, +41.4\% over proxy tuning across tasks). Through token-level analysis and physician case studies, we demonstrate that CAPT amplifies clinically actionable language, reduces context errors, and increases clinical specificity. This technique especially benefits healthcare institutions with constrained computational capacity that cannot support iterative clinical training and want to adopt emerging general-domain model advances.
\end{abstract}

\paragraph*{Data and Code Availability}
\paragraph*{Data and Code Availability}
All datasets used in this paper are publicly available and shown in Table~\ref{tab:tasks}. Code available at: \url{https://github.com/sronaghi/training_free_adaptation}.


\paragraph*{Institutional Review Board (IRB)}
This research uses publicly available datasets and does not require IRB approval.

\section{Introduction}
\label{sec:intro}
Despite advances in general-domain language models, their application to clinical settings remains limited by hallucinations, omission of critical details, and failures in clinical reasoning \citep{lehman2023dostillneed,hager2024llm_clinical_limits,asgari2025safety_llms}. These shortcomings arise because pretraining corpora contain limited representations of clinical data, such as electronic health records, due to privacy constraints \citep{singhal2023nature_clinical_knowledge}. Moreover, reliance on large-scale Internet text during training encodes biases, outdated information, or incorrect medical knowledge \citep{Alber2025Poisoning, Zack2024GPT4Bias, wulimitationsfinetune}.

Domain adaptation techniques such as continued pretraining and instruction tuning address these challenges but require resource-intensive re-training for each new model generation (e.g., MedPalm 1 $\rightarrow$ MedPalm 2) \citep{singhal2025naturemed_mqa}. This creates a lag between advances in base model capabilities and their clinical applicability, with substantial computational resources reinvested in domain adaptation for each new architecture. As a result, on comprehensive clinical benchmarks, general-domain state-of-the-art models often outperform clinically adapted models \citep{wu2025bridgebenchmarkinglargelanguage}. 

\begin{figure*}[t]
    \centering
    \includegraphics[width=\textwidth]{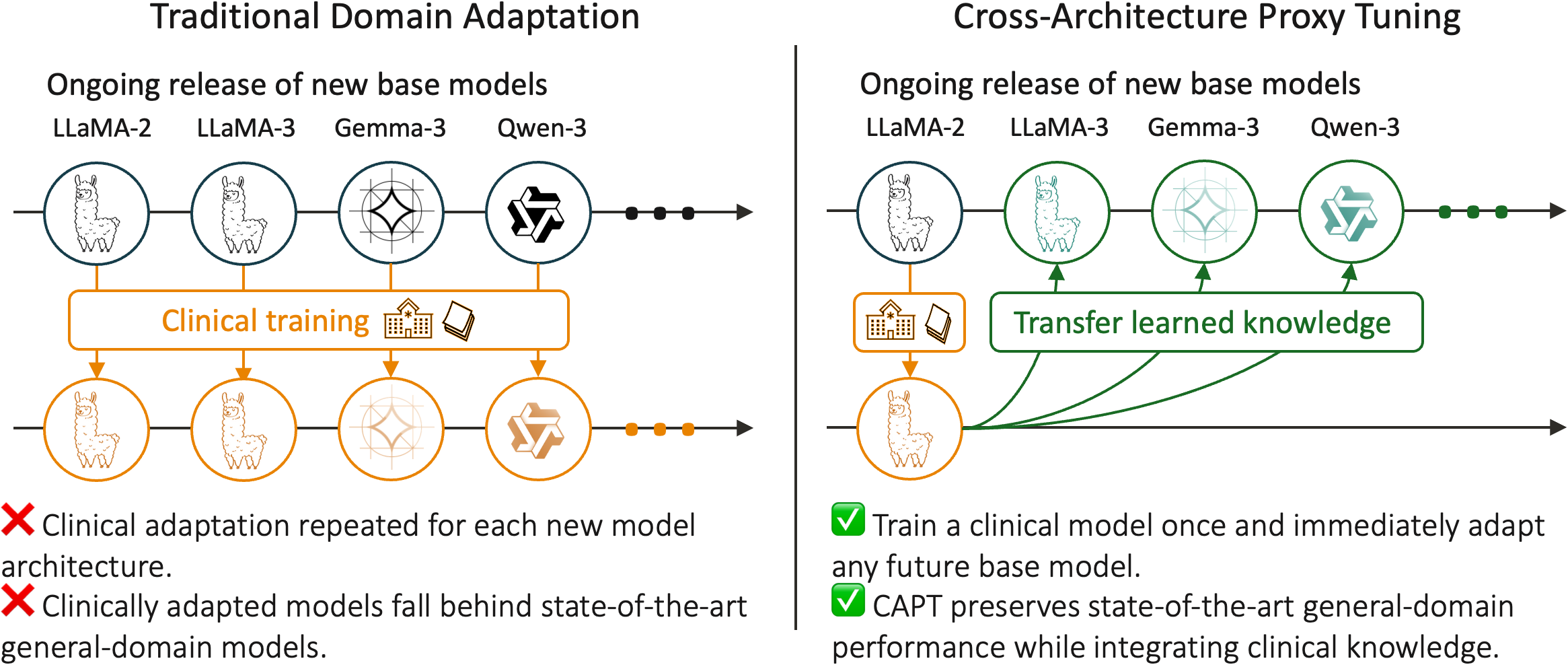}

    \caption{\textbf{Cross-Architecture Proxy Tuning (CAPT) Overview:} Conventional approaches require costly retraining for each new architecture (left), causing clinical models to lag behind general-domain counterparts. CAPT transfers knowledge from a single clinical model to new-generation models without additional training (right).}
    \label{fig:overview}
\end{figure*}

We explore model ensembling as a training-free approach to combine the advanced reasoning of new-generation general-domain models with the domain knowledge of legacy clinical models. Prior works show that integrating decoding-time probability distributions of general-domain models with heterogeneous architectures yields efficient performance gains \citep{yao2025determinethenensemblenecessitytopkunion}. However, these approaches assume largely overlapping model capabilities and have not been explored for combining models with distinct strengths, where information must be selectively integrated rather than uniformly aggregated. For example, we seek to incorporate the learned clinical knowledge of an older-generation clinical model without inheriting its limitations in instruction following, reasoning, or robustness due to its architecture or catastrophic forgetting \citep{Kirkpatrick_2017}.

Contrastive decoding offers a mechanism for isolating a model’s strengths by choosing tokens that maximize the likelihood difference between expert and amateur models, amplifying the expert's behavior \citep{li-etal-2023-contrastive}. Proxy tuning leverages contrastive decoding to efficiently tune a large, pretrained model by combining its logit distribution with the distributional delta between a smaller fine-tuned model and its untuned, base counterpart \citep{liu2024proxy}. However, proxy tuning requires shared tokenization across models, restricting architectural diversity. This constraint prevents reuse of domain-adapted models built on older architectures, restricting adoption in resource-limited settings.

Here, we propose \textbf{Cross-Architecture Proxy Tuning (CAPT)}, a probability-level ensembling method that supports models with disjoint vocabularies and leverages contrastive decoding to selectively incorporate the specialized knowledge of a domain-adapted model (Figure~\ref{fig:overview}). Our contributions are three-fold: \begin{itemize}
    \item  We introduce CAPT which enables reuse of legacy domain-adapted clinical models for training-free adaptation of newer-generation models.  On six clinical classification and text-generation benchmarks, CAPT consistently outperforms prior ensembling approaches, with an average improvement of 17.6\% over UniTE and 41.4\% over proxy tuning across metrics.
    \item Through a token-level analysis, we illustrate that CAPT selectively integrates the clinical model's knowledge for tokens related to clinical decision-making and documentation style, while the general model controls linguistic structure and formatting tokens. 
    \item In case studies by two board-certified physicians, we demonstrate that CAPT-generated outputs contain more precise clinical terminology, context-appropriate recommendations, and improved clinical accuracy.
\end{itemize}
\section{Related Works} 
\xhdr{Clinical Domain Adaptation} Clinical models such as Gatortron, clinicalBERT, NYUTron, Me-LLaMA, and Clinical-T5 have demonstrated the effectiveness of continued pretraining on large volumes of unlabeled medical text \citep{yang2022gatortronlargeclinicallanguage, alsentzer2019publiclyavailableclinicalbert, jiang2023healthsystem, xie2024mellama, lehman2023clinicalt5}. Following continued pretraining, supervised fine-tuning can further improve performance on specific clinical tasks, instruction following, and alignment with human preferences \citep{han2023medalpaca, zhang2023alpacare, singhal2023nature_clinical_knowledge, wang2024clinicnote, zhang2023huatuogpt}. While effective, these approaches are highly resource-intensive, especially continued pre-training; for example, Me-LLaMA conducted continued pretraining on LLaMA-2 base models using 129B tokens and 160 × 80 GB A100 GPUs \citep{xie2024mellama}. Our work complements these techniques by addressing their limited transferability, as these costly adaptations must be repeated for each new base model.

\xhdr{Model Ensembling} Probability-level model ensembling methods for heterogeneous architectures have primarily been explored with multiple general-domain models. Existing works focus on optimizing efficiency and vocabulary alignment: EVA learns mappings between LLM vocabularies through their existing vocabulary overlap \citep{xu2024bridginggapdifferentvocabularies}, and DeePEn projects the probability distributions of multiple models onto a unified space \citep{huang2024ensemblelearningheterogeneouslarge}. UNiTE, the strongest-performing method to date, unions each model’s top-$k$ tokens at every decoding step and combines probabilities via re-tokenization \citep{yao2025determinethenensemblenecessitytopkunion}. Prior analyses show ensembling methods only outperform the ensemble's strongest model when performance gaps are within 10\% \citep{yao2025determinethenensemblenecessitytopkunion}. This limits applicability when combining new- and old-generation architectures where gaps can arise from architectural advances alone \citep{bedi2025medhelm}. In contrast, CAPT is designed for models with asymmetric capabilities and enables selective integration of learned information from older domain models.

\xhdr{Contrastive Decoding}
Contrastive decoding selects tokens that are more likely under a strong language model by contrasting its logit or probability output distribution against that of a weaker model. Prior work operationalizes this idea by combining the logits of a large model with the distributional delta between a smaller strong model and its untuned base counterpart, steering generation toward desirable behaviors such as reduced toxicity \citep{liu-etal-2021-dexperts}, increased helpfulness \citep{mitchell2023eft}, selective unlearning \citep{suriyakumar2025ucdunlearningllmscontrastive}, and improved coding, question answering, and mathematical reasoning \citep{liu2024proxy}. These approaches typically assume shared vocabularies and are evaluated in task- or instruction-tuning regimes that remain in-distribution with respect to the large model’s pretraining data. In contrast, our approach supports models with disjoint vocabularies and is evaluated on clinical domain adaptation, a substantially out-of-distribution setting \citep{Kim_2025}.

\section{Methods}
\subsection{Cross-Architecture Proxy Tuning}
\xhdr{Overview}
We hypothesize that general-domain models encode substantial medical knowledge but lack exposure to clinical practice patterns and stylistic conventions reflected in clinical notes. Accordingly, the new-generation general-domain model should lead generation to preserve fluency and reasoning, while the clinical model selectively interjects for tokens associated with domain-specific reasoning or stylistic patterns. To achieve this, CAPT re-ranks the top-$k$ candidate tokens proposed by the general-domain model using a log-probability offset defined by the difference between the clinically trained model and its untrained base counterpart.

At each decoding step, we select the top-k next-token candidates from the new-generation general-domain model. Each candidate token is re-tokenized using the clinical model’s tokenizer to obtain a corresponding clinical token. We then adjust the candidate’s log-probability by adding the log-probability difference between the clinical model and its base counterpart. The highest-scoring token after adjustment is appended to the context.

\begin{figure}[t]
    \centering
    \includegraphics[width=\linewidth]{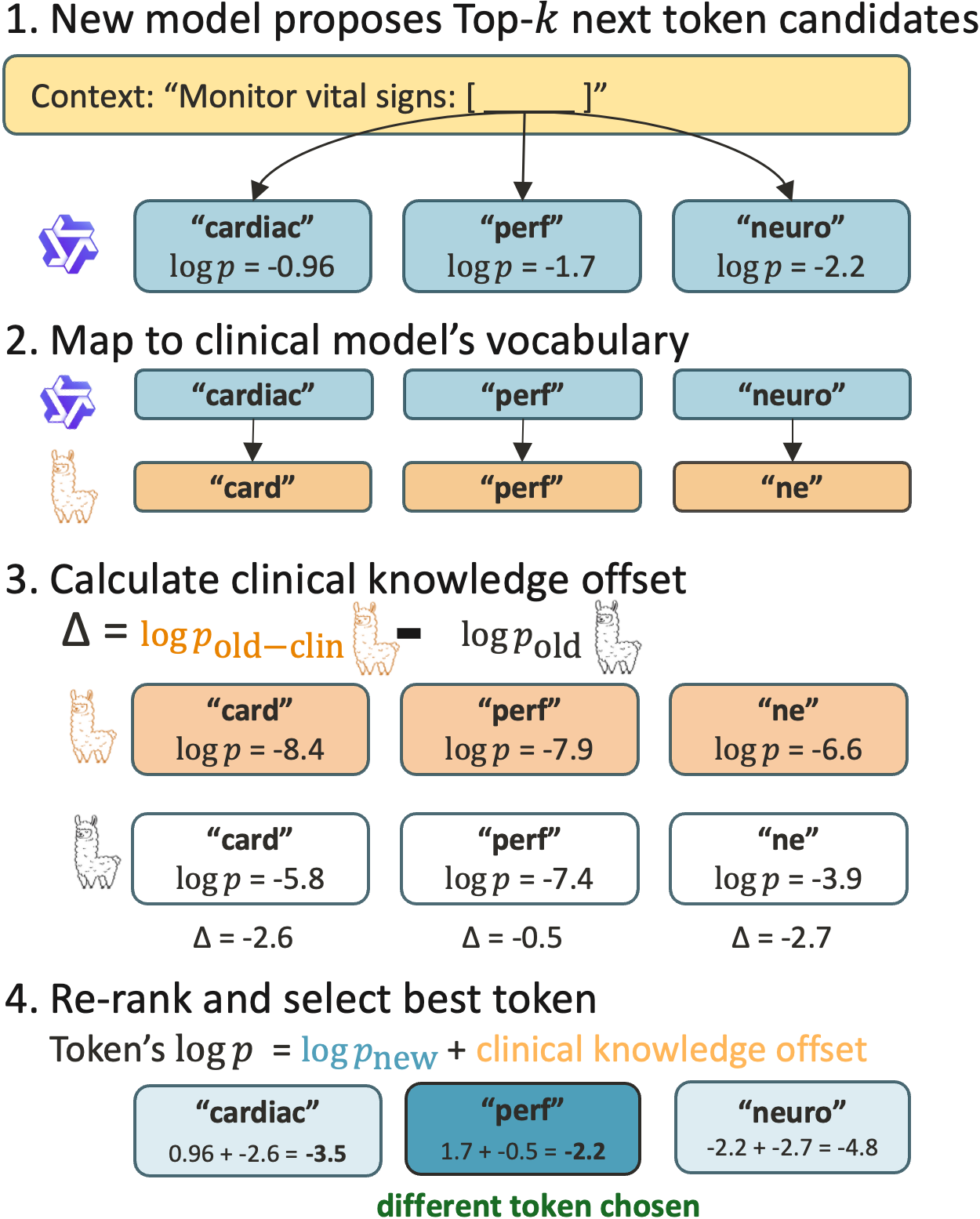}
    \caption{\textbf{CAPT decoding process.} At each step, CAPT (1) selects top-$k$ token candidates from the new-generation general domain model, (2) maps each token candidate to the clinical model's vocabulary via retokenization, (3) computes the clinical knowledge offset as the log-probability difference between the clinical model and its untuned counterpart, and (4) re-ranks candidates by adding the clinical offset to the original log-probabilities. In this example, CAPT selects ``perf" (perfusion) over ``cardiac" for monitoring a forearm graft, as the clinical offset adjusts the ranking to favor the contextually appropriate term. Full example context can be found in Figure~\ref{fig:example}a.}
    \label{fig:token_categories}
\end{figure}

\xhdr{Formal definition}
Let $M_{\text{new}}$ be a new-generation general-domain language model with tokenizer $\mathcal{T}_{\text{new}}$ and vocabulary $\mathcal{V}_{\text{new}}$.
Let $M_{\text{old-clin}}$ denote an older-generation clinically trained model and
$M_{\text{old}}$ its untrained base counterpart; both share tokenizer
$\mathcal{T}_{\text{old}}$ and vocabulary $\mathcal{V}_{\text{old}}$.

Given a context $\mathbf{x}_{1:t}$, the models define next-token log-probabilities
$\log p_{\text{new}}(\cdot\mid \mathbf{x}_{1:t})$ over $\mathcal{V}_{\text{new}}$ and
$\log p_{\text{old-clin}}(\cdot\mid \mathbf{x}_{1:t})$, $\log p_{\text{old}}(\cdot\mid \mathbf{x}_{1:t})$ over $\mathcal{V}_{\text{old}}$.
We restrict computation to the candidate set
$\mathcal{C}_t = \text{Top-$k$}\big(\mkern-1mu\log p_{\text{new}}(\cdot\mid \mathbf{x}_{1:t})\big) \subseteq \mathcal{V}_{\text{new}}$, the $k$ most likely next-tokens under $M_{\text{new}}$.

Because the models operate over different vocabularies, we define a decode--retokenize mapping
$f : \mathcal{V}_{\text{new}} \rightarrow \mathcal{V}_{\text{old}}$
that projects each candidate token from the new model into the clinical
model’s vocabulary by decoding a token from $\mathcal{V}_{\text{new}}$ to its string form
and re-tokenizing the string using $\mathcal{T}_{\text{old}}$.
For $i \in \mathcal{C}_t$, we define $f(i)$ as the first non-space token in
$\mathcal{T}_{\text{old}}(\mathrm{decode}_{\text{new}}(i))$.
For each candidate token $i \in \mathcal{C}_t$, we update
its log-probability:
\begin{multline*}
s(i) = \log p_{\text{new}}(i \mid \mathbf{x}_{1:t})\; +\\
\quad\alpha\Big(\!\log p_{\text{old-clin}}(f(i)\mid \mathbf{x}_{1:t}) - \log p_{\text{old}}(f(i)\mid \mathbf{x}_{1:t})\!\Big)
\end{multline*}
which adds a correction representing the isolated clinical-domain signal. The selected token, $i^*=\arg\max_{i \in \mathcal{C}_t} s(i)$, is appended to the context. 
We set $k = 20$, reflecting the top-20 log-probabilities exposed by black-box API providers \citep{openai2025gpt5, google2025gemini3} and fix  
$\alpha = 1.0$. See Appendix~\ref{sec:hyperparameter} for hyperparameter analysis.

\begin{table*}[t]
\centering
\setlength{\tabcolsep}{6pt}
\caption{\textbf{Evaluation Tasks}}
\vspace{2pt}
\label{tab:tasks}
\begin{tabular}{
    p{3.0cm}   
    p{7.0cm}   
    p{5.2cm}   
}
\toprule
\textbf{Task} & \textbf{Input} & \textbf{Output} \\
\midrule
\multicolumn{3}{l}{\textbf{Classification}} \\

MedNLI &
Premise from clinical note + hypothesis &
Entailment/Neutral/Contradiction \\

MTS-Specialty &
Medical transcription note &
One of 40 medical specialties \\

CLIP &
Discharge summary sentence &
One or more of 7 follow-up labels \\
\midrule
\multicolumn{3}{l}{\textbf{Text Generation}} \\

MIMIC-RRS &
Radiology report findings section &
Impression section \\

MIMIC-BHC &
Full discharge summary note &
Brief Hospital Course summary \\

MTS-Procedure &
Operative note or surgical case description &
Postoperative treatment plan \\
\bottomrule
\end{tabular}
\end{table*}

\subsection{Models and Baselines} We use Qwen3-30B \citep{yang2025qwen3technicalreport} as the new-generation general-domain model ($M_{\text{new}}$) and MeLLaMA-13B-chat \citep{xie2024mellama} as the old-generation clinical model ($M_{\text{old-clin}}$). MeLLaMA-13B-chat is continually pre-trained from LLaMA-2-13B-base ($M_{\text{old}}$) and instruction-tuned on clinical notes and tasks. We compare CAPT to proxy tuning \citep{liu2024proxy} and UNiTE \citep{yao2025determinethenensemblenecessitytopkunion}, the strongest probability-level ensembling method to date. Since proxy tuning requires a shared tokenizer between models, we use LLaMA-2-70B-chat ($M_{\text{old-L}}$) \citep{touvron2023llama2openfoundation} as its general-domain model. 

\subsection{Evaluation Tasks} As shown in Table~\ref{tab:tasks}, we evaluate CAPT on six clinical classification and text-generation tasks to assess performance across a broad range of use cases. 
For classification tasks, we report Macro-F1 and accuracy on 200 random test samples from:
\textit{MedNLI} \citep{romanov2018lessonsnaturallanguageinference}, \textit{MTS-Specialty} \citep{MTSamples}, and
\textit{CLIP} \citep{mullenbach2021clipdatasetextractingaction}.

For text generation, we report MedHELM LLM-jury scores \citep{bedi2025medhelm} and MedVAL percent of risk-free outputs \citep{aali2025medvalexpertlevelmedicaltext} on 100 samples from: \textit{MIMIC-BHC} \citep{Aali_2024}, \textit{MTS-Procedures} \citep{MTSamples}, and \textit{MIMIC-RRS} \citep{Chen_2023}.

MedHELM's LLM-jury metric evaluates accuracy, completeness, and clarity of generated outputs against reference outputs. Our results can be compared to MedHELM's leaderboard, and CAPT can be readily applied to any of the 37 MedHELM clinical classification, text generation, or question-answering tasks. MedVAL is a fine-tuned LLM that assigns a risk score to outputs, where level 1 is risk-free and level 4 is the highest risk level. Both metrics are validated against clinician judgement. See Appendix~\ref{sec:tasks} for details on experimental setup and task prompts. 

\subsection{Token-level Analysis}
To understand how CAPT leads to token-level changes, we manually categorized 280 of the top 25\% most frequently generated tokens in the MTS-Procedure task into semantic categories. The remaining tokens were excluded because they were ambiguous, subword fragments, or otherwise not semantically interpretable. We use MTS-Procedure because it is publicly available; our other text-generation tasks rely on MIMIC \citep{johnson2016mimic}, whose raw data cannot be shared. See Appendix~\ref{sec:token_categories} for token categories and corresponding tokens.  

For each category, we report the mean per-token log-probability difference between the clinical model $M_{\text{old-clin}}$ and its base counterpart $M_{\text{old}}$.  Because shifts are computed over generated tokens, \sethlcolor{cborange!80}\hl{positive} values indicate tokens preferentially up-weighted by the \sethlcolor{cborange!80}\hl{clinical} model, while  \sethlcolor{cbblue!60}\hl{negative} values indicate tokens supported by the new-generation \sethlcolor{cbblue!60}\hl{general-domain} model, such that they remain selected even when the clinical offset disfavors them. 
\subsection{Physician Case Study}
We present five randomly selected full-length CAPT outputs from the MTS-Procedure task, annotated with token-level shifts and expert qualitative analysis. The analysis was conducted by two board-certified physicians, an interventional radiologist with 30 years of clinical experience and an anesthesiologist with 5 years of clinical experience and an expertise in clinical language models. Similar to the token-level analysis, we use MTS-Procedure because it is publicly available, whereas our other text-generation tasks rely on MIMIC \citep{johnson2016mimic}. The physicians assessed clinical accuracy, appropriateness, and overall utility.

\section{Results} 
\label{sec:results}

\begin{table*}[htbp]
\centering
\caption{\textbf{Performance comparison across tasks and methods.} Bold indicates best performance. F1 = Macro-F1, Acc = Accuracy, LLM-J = MedHelm LLM-jury score (average out of 5), \%RF = MedVAL \% of risk-free outputs.}
\label{tab:results}
\footnotesize 
\resizebox{1.02\textwidth}{!}{
\begin{tabular}{lcccccccccccc}
\toprule
& 
\multicolumn{6}{c}{\textbf{Classification}} & 
\multicolumn{6}{c}{\textbf{Text-Generation}} \\
\cmidrule(lr){2-7} \cmidrule(lr){8-13}
& 
\multicolumn{2}{c}{\textbf{MedNLI}} & 
\multicolumn{2}{c}{\makecell{\textbf{MTS-Specialty}}} & 
\multicolumn{2}{c}{\textbf{CLIP}} & 
\multicolumn{2}{c}{\makecell{\textbf{MTS-Procedure}}} & 
\multicolumn{2}{c}{\textbf{MIMIC-RRS}} & 
\multicolumn{2}{c}{\textbf{MIMIC-BHC}} \\
\textbf{Method / Models} & F1 & Acc & F1 & Acc & F1 & Acc & LLM-J & \%RF & LLM-J & \%RF & LLM-J & \%RF \\
\midrule
\makecell[l]{\textbf{Old-gen large general} ($M_{\text{old-L}}$) \\LLaMA-2-70B-chat}
& 0.598 & 0.635 & 0.094 & 0.280 & 0.097 & 0.205 & 3.214 & 26.67 & 2.453 & 57.50 & 3.323 & 13.00 \\
\makecell[l]{\textbf{Old-gen clinical} ($M_{\text{old-clin}}$)\\MeLLaMA-13B-chat}
& 0.539 & 0.535 & 0.080 & 0.210 & 0.192 & 0.785 & 1.989 & 45.83 & 3.379 & 35.00 & 2.906 & 14.14 \\
\makecell[l]{\textbf{New-gen general} ($M_{\text{new}}$)\\Qwen3-30B}
& 0.858 & 0.850 & \textbf{0.140} & \textbf{0.350} & \textbf{0.240} & 0.540 & 3.825 & 63.33 & 4.394 & \textbf{65.00} & 3.932 & 30.00 \\
\makecell[l]{\textbf{Proxy Tuning}\\$M_{\text{old-L}}$ + $M_{\text{old-clin}}$ + $M_{\text{old}}$}
& 0.575 & 0.620 & 0.095 & 0.275 & 0.143 & 0.420 & 3.466 & 55.00 & 3.898 & 34.17 & 3.456 & 8.00 \\
\makecell[l]{\textbf{UniTE}\\$M_{\text{new}}$ + $M_{\text{old-clin}}$}
& 0.873 & 0.875 & 0.060 & 0.100 & 0.217 & 0.560 & 3.486 & 50.00 & 3.903 & 46.67 & 3.747 & 20.00 \\
\makecell[l]{\textbf{CAPT}\\ $M_{\text{new}}$ + $M_{\text{old-clin}}$ + $M_{\text{old}}$}
& \textbf{0.886} & \textbf{0.885} & 0.135 & 0.320 & 0.224 & \textbf{0.580} & \textbf{3.882} & \textbf{70.83} & \textbf{4.414} & 56.67 & \textbf{3.973} & \textbf{31.00} \\
\bottomrule
\end{tabular}
}
\end{table*}

\subsection{Model Performance} 
CAPT outperforms baselines and existing model-ensembling approaches. Table~\ref{tab:results} summarizes performance across clinical classification and text generation tasks. $M_{\text{new}}$ outperforms $M_{\text{old-clin}}$ by an average of 53.6\% in Macro-F1 on classification benchmarks and 52.6\% in LLM-jury scores on text-generation benchmarks. This performance gap likely explains why UNiTE, which averages the probability scores of $M_{\text{old-clin}}$ and $M_{\text{new}}$, performs worse than $M_{\text{new}}$ alone by 8–28\% on average. Proxy tuning improves upon both $M_{\text{old-L}}$ and $M_{\text{old-clin}}$ despite their large performance gap, demonstrating the potential of contrastive decoding-based methods to bridge models of distinct capabilities. However, $M_{\text{new}}$ consistently outperforms proxy tuning, highlighting how the method's reliance on shared vocabularies limits its applicability as new base models are released. 

CAPT outperforms $M_{\text{new}}$ on 8 of 12 metrics, while UniTE improves on only 2. Averaged across all tasks and metrics, CAPT outperforms UniTE and proxy tuning by 17.6\% and 41.4\%, respectively. CAPT's gains are particularly pronounced on text-generation tasks, where clinical reasoning and terminology precision are critical. For these tasks, CAPT yields substantially more risk-free outputs, improving by 13.94 and 20.44 percentage points over UniTE and proxy tuning respectively. This indicates meaningful improvements in clinical safety and utility alongside performance gains.

\begin{figure}[t]
    \centering
    \includegraphics[width=\linewidth]{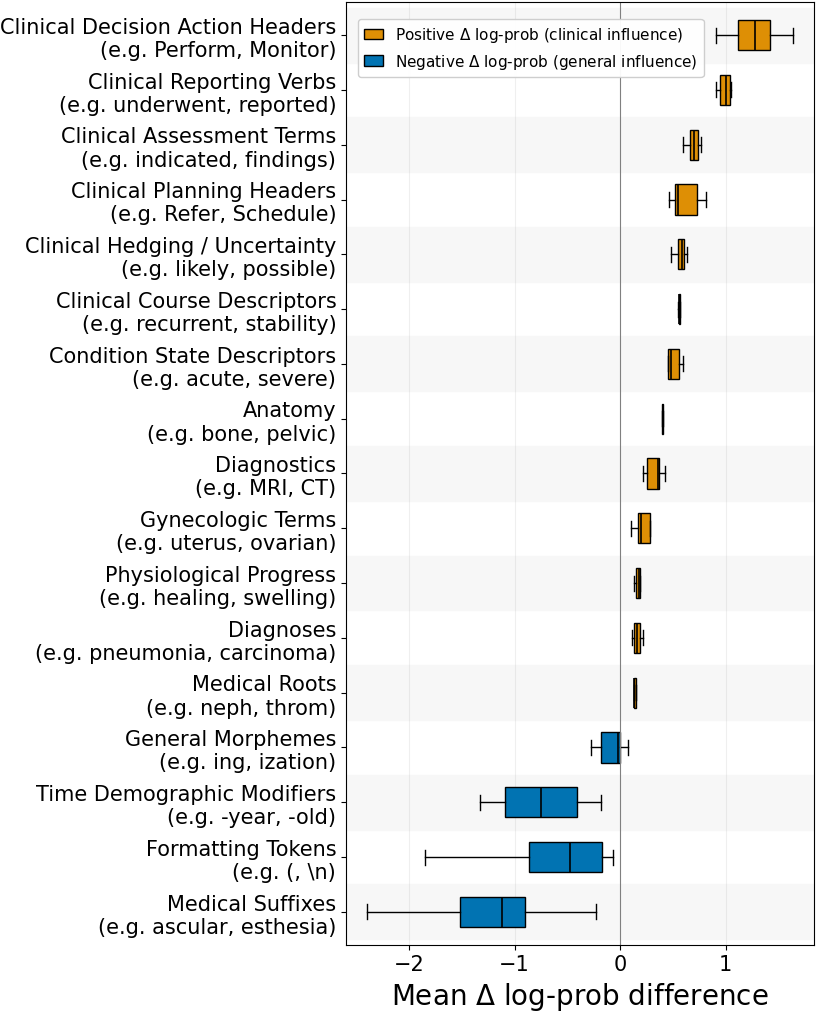}

    \caption{\textbf{Mean log-probability offset between $M_{\text{old-clin}}$ and $M_{\text{old}}$ of generated tokens by semantic category.} Positive shifts (\sethlcolor{cborange!80}\hl{orange}) indicate increased influence of $M_{\text{old-clin}}$, while negative shifts (\sethlcolor{cbblue!60}\hl{blue}) of $M_{\text{new}}$. All token categories are shown in Appendix~\ref{sec:token_categories}, Figure~\ref{fig:all_token_categories1}.}
    \label{fig:token_categories}
\end{figure}

\begin{figure*}[t]
    \centering \includegraphics[width=\textwidth]{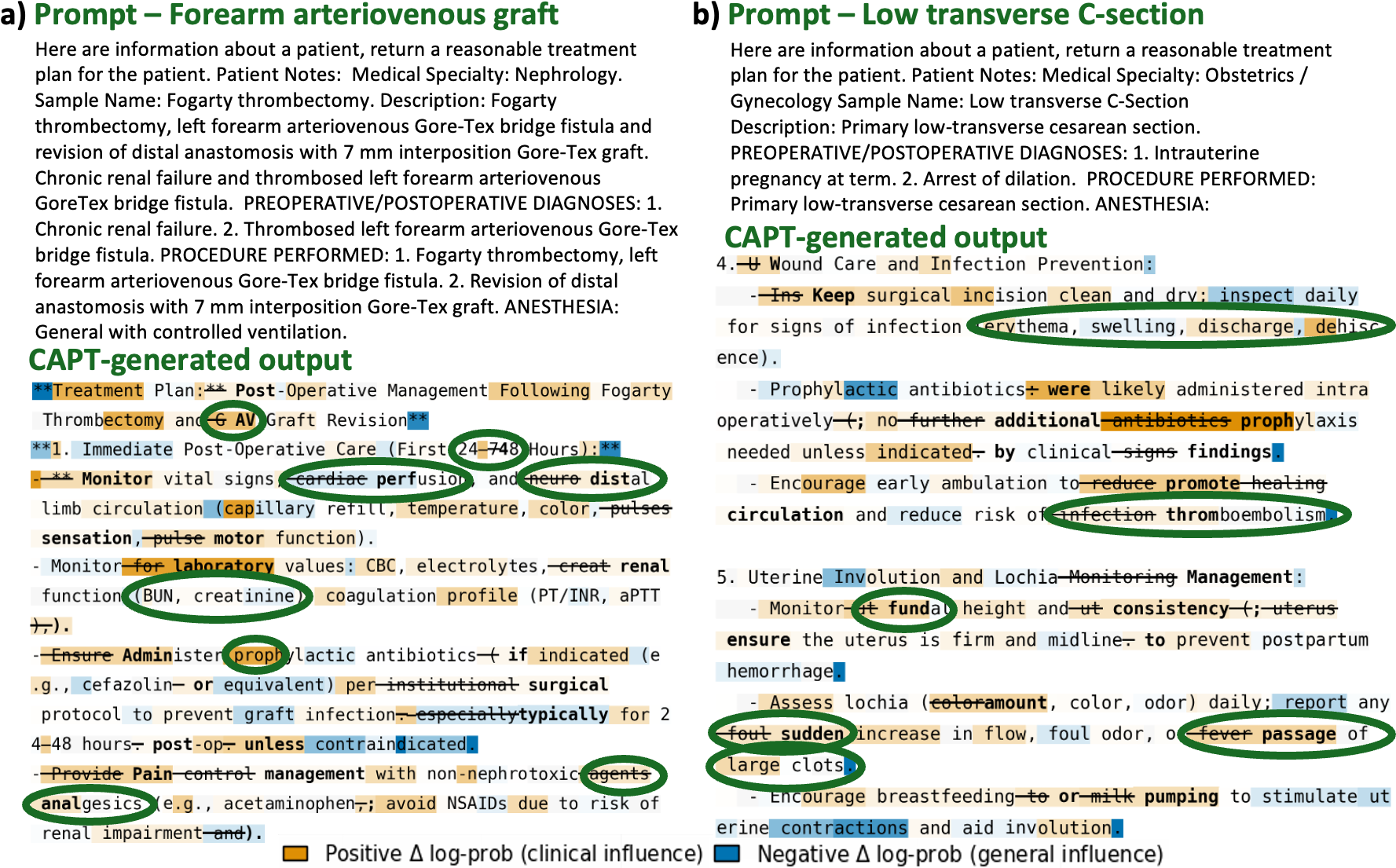}
    \caption{\textbf{CAPT-generated post-operative management plans} for two MTS-Procedure examples. \sethlcolor{cborange!80}\hl{Orange} and \sethlcolor{cbblue!60}\hl{blue} highlights indicate stronger influence from $M_{\text{old-clin}}$ and $M_{\text{new}}$, respectively. \textbf{Bolded} tokens denote when the top-choice token changed after CAPT adjustment, with the original top-choice \st{crossed out}. \textcolor{ForestGreen}{Green} circles mark tokens discussed in main text. The full outputs, extended analysis, LLM jury evaluation, and additional case studies are shown in Appendix~\ref{sec:case_study}.}
    \label{fig:example}
\end{figure*} 

\subsection{Token-level Analysis} 
CAPT selectively integrates clinical domain knowledge. Figure~\ref{fig:token_categories} summarizes mean log-probability shifts by semantic token category. Positive values indicate categories for which CAPT increases token preference via the clinical-model offset between $M_{\text{old-clin}}$ and $M_{\text{old}}$, while negative values indicate categories for which the offset decreases preference and $M_{\text{new}}$ retains stronger influence. 

$M_{\text{old-clin}}$ strongly influences tokens associated with clinical decision-making (e.g., \textit{Clinical Decision Action Headers, Clinical Assessment Terms}) and documentation style (e.g., \textit{Clinical Reporting Verbs, Clinical Hedging, Condition State Descriptors}). These positive shifts reflect the clinical model's learned conventions for structuring clinical assessments, expressing uncertainty, and framing care decisions. These stylistic patterns are largely absent from general-domain pretraining corpora, which typically exclude clinical notes, indicating that CAPT effectively transfers knowledge learned by an older-generation model. 

In contrast, $M_{\text{new}}$ dominates linguistic structure and formatting (e.g., \textit{Formatting Tokens, General Morphemes, Medical Suffixes}), reflecting that the general-domain model controls grammatical coherence and document structure. Similarly, \textit{Time Demographic Modifiers} show negative shifts, indicating that $M_{\text{new}}$ influences temporal precision and reasoning. These patterns demonstrate that CAPT preserves the newer model's capabilities in general reasoning and fluency.

Notably, medical knowledge categories exhibit near-zero shifts, indicating both models converge on similar preferences for foundational medical concepts. Core medical terminology including \textit{Gynecologic Terms, Diagnoses, Medical Roots}, and \textit{Physiological Progress} all cluster near zero. This suggests CAPT primarily affects how $M_{\text{new}}$'s medical knowledge is expressed to match clinical documentation conventions.

\subsection{Physician Case Study} Figure~\ref{fig:example} shows samples from two case studies generating postoperative treatment plans for the MTSamples-Procedure task. See Appendix~\ref{sec:case_study} for extended analysis, LLM jury evaluation, and additional examples. 
\subsubsection{Case Study 1}
Figure~\ref{fig:example}a shows a sample of a CAPT-generated treatment plan for a forearm arteriovenous graft procedure (graft connecting vein to artery). 

CAPT token-level shifts improve clinical accuracy. CAPT replaces ``72'' with ``48'' in the postoperative monitoring timeline, narrowing the range from 24-72 hours to 24-48 hours in line with typical discharge practices for this procedure. Similarly, $M_{\text{new}}$ recommends monitoring ``cardiac'' symptoms, which are not relevant to a forearm graft; CAPT correctly emphasizes perfusion (i.e. blood flow). 

CAPT also shifts generation toward more precise clinical terminology, resulting in a more clinically actionable and descriptive plan. CAPT replaces $M_{\text{new}}$'s ``neuro'' token, which likely implies ``neuro check,'' with ``dist.'' This change directs generation towards monitoring ``distal limb circulation,'' which encompasses relevant components of the neurologic assessment (e.g., sensation and motor function) as well as other important checks (e.g., capillary refill). Similarly, CAPT inserts ``AV'' before ``graft'' to clarify the graft type and replaces  ``agents'' with ``analgesics'' to specify pain medication rather than any generic medication class. CAPT also heavily preferences ``proph'' for ``prophylactic antibiotics'' after ``Administer,'' specifying the preventative nature of the antibiotic administration. 

One limitation is that CAPT preserved over-explanatory content intended for non-expert audiences (e.g., naming specific antibiotics), which are unnecessary for nurses responsible for post-operative management. These additions often appear in parentheses (e.g., ``('') that are favored by $M_{\text{new}}$, reflecting its general-audience training objective.

\subsubsection{Case Study 2}
Figure~\ref{fig:example}b shows a CAPT-generated treatment plan for a low transverse cesarean section (C-section), a surgical procedure in which the fetus is delivered through a horizontal lower uterine incision.

CAPT increases procedure-specific, safety-critical recommendations. The clinical model increases preference towards ``erythema, discharge, and dehiscence,'' procedure-specific signs of infection critical to monitor in C-sections. CAPT also adds ``fundal'' height, a highly procedure-specific physical exam sign that serves as a proxy for adequate uterine tone (contraction of the uterus, which is essential for hemorrhage prevention). CAPT replaces ``foul'' with ``sudden'' when describing changes in lochia (normal bloody discharge following delivery), reflecting a shift from monitoring general infection signs to detecting postpartum hemorrhage. Sudden increases in lochia  are a sign of delayed postpartum hemorrhage, whereas foul odor is a sign of infection or chorioamnionitis, a more rare complication after delivery than postpartum hemorrhage. Similarly, CAPT replaces monitoring for ``fever'' with ``passage of large clots,'' another sign of postpartum hemorrhage. Overall, this shift in focus towards postpartum hemorrhage is clinically appropriate, as hemorrhage is the leading cause of maternal morbidity worldwide and undergoing a C-section increases a patient's risk \citep{Yunas2025PPHMetaAnalysis}. 

CAPT additionally reduces clinical errors. In the recommendation to encourage early ambulation (moving during recovery for circulation), CAPT replaces ``infection'' with ``thromboembolism'' (blood clot forming and traveling to block blood flow). Early ambulation reduces risk of thromboembolism, but does not reduce risk of infection. 

\subsubsection{Takeaways}
Across both cases, CAPT makes three types of improvements: (1) \textit{increased clinical specificity} (replacing generic term ``agents'' with more precise ``analgesics''), (2) \textit{enhanced clinical and contextual accuracy} (``24-72''$\rightarrow$ ``24-48'' for postoperative monitoring timeline; ``infection''$\rightarrow$``thromboembolism'' for risks reduced with ambulation), and (3) \textit{emphasizing procedure-specific, safety-critical monitoring} (``distal limb circulation'' for a lower forearm graft; postpartum hemorrhage signs for a C-section). We find that CAPT preserves $M_{\text{new}}$'s explanatory style, which may be redundant for expert clinicians. Although a limitation, this suggests that CAPT incorporates clinical expertise without altering the base model's communication patterns.

\section{Discussion}
We introduce \textit{Cross-Architecture Proxy Tuning} (CAPT), a training-free model-ensembling strategy that adapts newly released general-domain language models to clinical use by reusing legacy clinical models. CAPT targets a practical bottleneck in clinical NLP deployment: continued pretraining and instruction tuning must be repeated for each new model generation, which is often infeasible for resource-constrained health systems. By operating at decoding time and supporting disjoint tokenizers, CAPT provides a path to benefit from new architecture advances without reinvesting in large-scale clinical retraining.

Across six clinical classification and text-generation tasks, CAPT consistently improves over strong baselines and prior heterogeneous ensembling methods, including cases where naive averaging harms performance due to large capability gaps between models. 
Our mechanistic analyses offer evidence for how CAPT achieves these gains. Token-level shifts indicate that the clinical model primarily influences decision-relevant phrasing and documentation conventions, while the new-generation model retains control over linguistic structure and global coherence. This division of labor is consistent with the intended design of CAPT: rather than transferring broad medical knowledge, CAPT preferentially alters how medical knowledge is expressed so that outputs better match clinical note style and clinical action framing. Physician case studies support this interpretation, showing that CAPT promotes  reduces context errors that can affect downstream usability.

This study has several limitations that motivate future work. First, clinically meaningful improvements can be sparse and context-dependent, and therefore are not always well captured by quantitative metrics. While we partially address this gap via physician review of CAPT outputs, robust evaluation of open-ended clinical generation remains an open challenge, particularly for measuring rare but safety-critical failures and subtle improvements in contextual appropriateness. Second, CAPT can inherit communication patterns from the general-domain model, including over-explanatory content that may be unnecessary for expert clinical audiences. Adaptively scaling the clinical-model adjustment coefficient (i.e. $\alpha$) based on intended user expertise could address this limitation. Third, CAPT increases inference-time compute by requiring forward passes through three models. Although we did not observe a significant wall-clock latency difference between Qwen3-30b alone and CAPT (which additionally includes MeLLaMA-13b-chat and LLaMA-2-13-base), the additional memory requirements may matter in deployment settings with stricter constraints. In this study, we use MeLLaMA-13B models, which are the smallest models in the MeLLaMA family and are state-of-the-art among open-source clinical language models. Exploring CAPT with smaller or more specialized clinical models may further improve the feasibility of this approach. 

Together, these results position CAPT as a practical bridge between rapidly advancing general-domain model architectures and slower-moving clinical adaptation pipelines. CAPT enables health systems and clinical NLP teams to reuse prior investments in domain-adapted models to make new-generation models more clinically aligned without additional training. More broadly, this work suggests a deployment strategy where legacy clinical models serve as reusable, modular adapters that can be paired with future foundation models, accelerating translation of new capabilities into clinical workflows while keeping adaptation costs and governance burdens manageable.


\bibliography{chil-sample}

\appendix \section{Experimental Setup}
\label{sec:tasks}
For classification tasks, we use prefix-constrained generation to constrain outputs to an explicit JSON schema consisting of a ``reason'' field (a free-text string with a maximum length of 600 characters) and a ``label'' field (a string or array of strings drawn from a predefined, task-specific label set). At each decoding step, the current prefix is used to determine the set of tokens that can legally follow while still permitting completion of a valid JSON object; all other tokens are masked out. In CAPT, the constraint is applied to the base model’s token distribution, since the base model alone determines the generated tokens. In proxy tuning and UNiTE, the constraint is applied to the token candidates considered during model combination, ensuring that only schema-valid tokens remain eligible after score aggregation. We employ constrained decoding to control for performance differences attributable to schema-adherence during instruction following.

Table \ref{tab:tasks_prompts} contains a description of each task and example prompt. All models were 4-bit quantized to fit on NVIDIA A100 40GB GPUs, such that our CAPT setup could run on a single 40GB GPU.

\section{Hyperparameter Analysis}
\label{sec:hyperparameter}
To evaluate the effect of $k$ and $\alpha$, we conducted experiments on two held-out, validation datasets. We report Macro-F1 on the validation split of the \emph{MedNLI} classification task, and MedHelm LLM-jury score on \emph{ACI-Bench}, a text-generation task of generating a structured clinical note from patient-doctor dialogues \citep{yim2023acibenchnovelambientclinical}. We use \emph{ACI-Bench} because MedHelm does not provide a validation set for its text-generation tasks and we did not want to reduce sample size of our tasks.
Table~\ref{tab:k_sensitivity} shows that varying $k$ has minimal impact on performance. We select $k=20$ because black-box API providers provide the top-20 logprobs for each token generation \citep{openai2025gpt5, google2025gemini3}. We set $\alpha = 1.0$ because it achieves best performance on both tasks, as shown by Table~\ref{tab:alpha_sensitivity}.

\begin{table}[H]
\centering
\resizebox{\columnwidth}{!}{
\begin{tabular}{lcc}
\toprule
$k$ & MedNLI (Macro-F1) & ACI-Bench (LLM-jury score) \\
\midrule
5  & 0.8944 & 4.377 \\
10 & 0.8944 & \textbf{4.380} \\
15 & 0.8944 & 4.377 \\
20 & \textbf{0.8992} & 4.379 \\
\bottomrule
\end{tabular}
}
\caption{Top-$k$ parameter experiments.}
\label{tab:k_sensitivity}
\end{table}

\begin{table}[H]
\centering
\resizebox{\columnwidth}{!}{
\begin{tabular}{lcc}
\toprule

$\alpha$ & MedNLI (Macro-F1) & ACI-Bench (LLM-jury score) \\
\midrule
0.5 & 0.8486 & 4.360 \\
0.7 & 0.8537 & 4.379 \\
1.0 & \textbf{0.8619} & \textbf{4.621} \\
\bottomrule
\end{tabular}}
\caption{$\alpha$ parameter analysis.}
\label{tab:alpha_sensitivity}
\end{table}

\section{Token-Level Analysis}\label{sec:token_categories}
\begin{figure*}[htp]
    \centering
    \includegraphics[width=.7\textwidth]{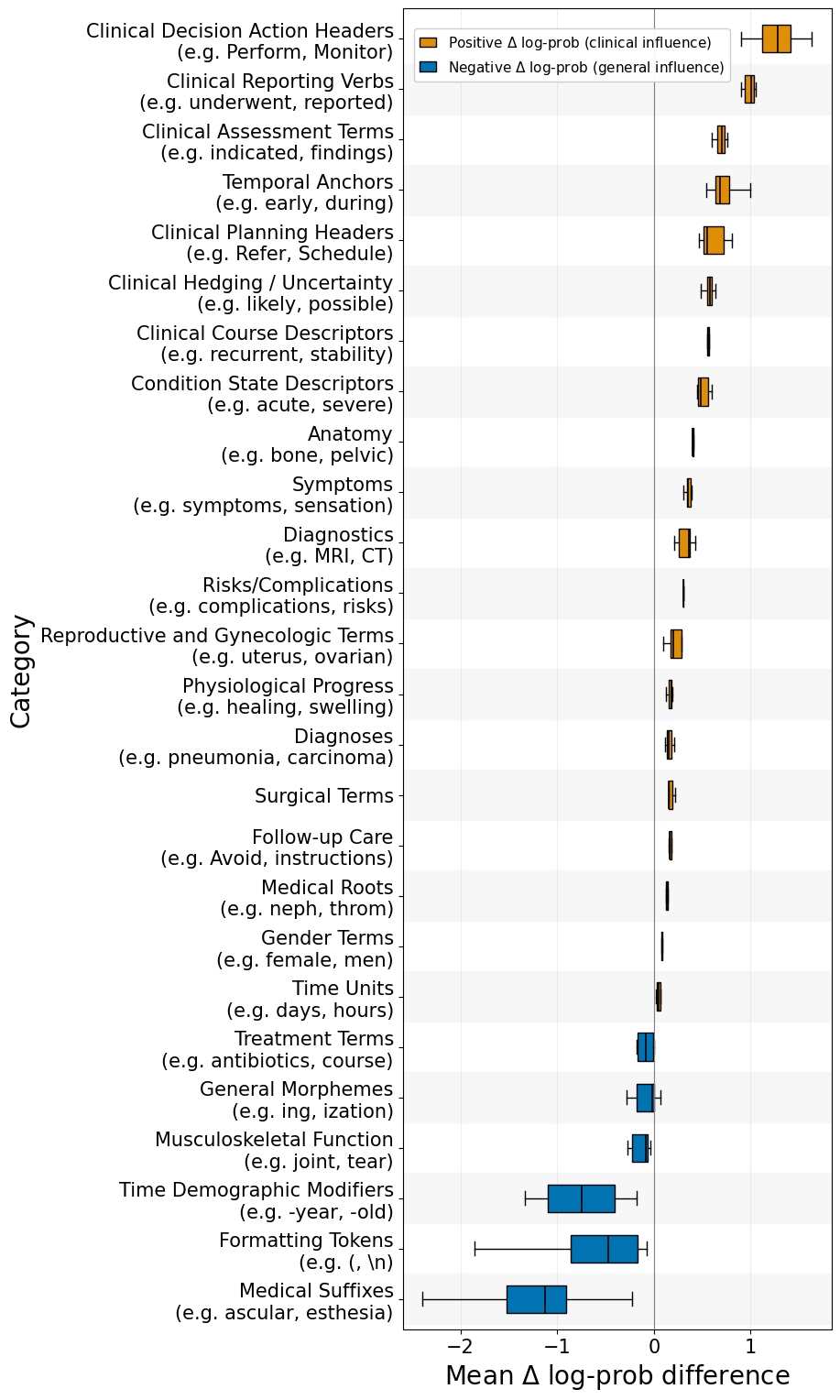}
    \caption{\textbf{Full version of Figure~\ref{fig:token_categories} with all token categories shown.} Mean log-probability offset between $M_{\text{old-clin}}$ and $M_{\text{old}}$ of generated tokens by semantic category. Positive shifts (\sethlcolor{cborange!80}\hl{orange}) indicate increased influence of $M_{\text{old-clin}}$, while negative shifts (\sethlcolor{cbblue!60}\hl{blue}) of $M_{\text{new}}$.}
    \label{fig:all_token_categories1}
\end{figure*}

Tables~\ref{tab:token_categories_w_tokens} and~\ref{tab:token_categories_cont}
show token categories and corresponding tokens. Figure~\ref{fig:all_token_categories1} shows the full version of Figure~\ref{fig:token_categories}, with all token categories shown. Due space constraints, Figure~\ref{fig:token_categories} only includes the categories discussed in the main text. 

\section{Physician Case Study}
\label{sec:case_study}
\begin{figure}[H]
\includegraphics[width=0.87\textwidth]{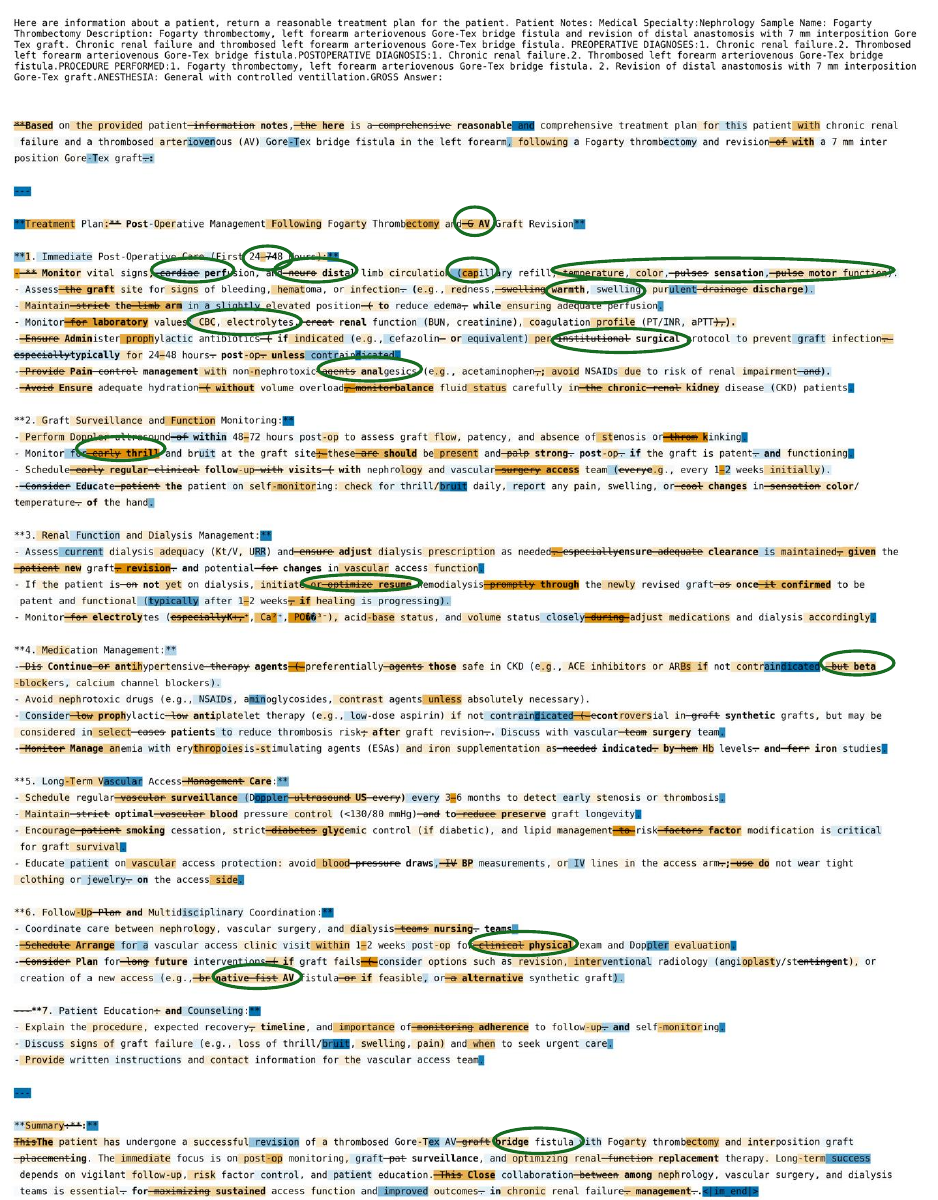}
\caption{\textbf{Case Study 1 - CAPT Output, LLM Jury Score = 4.22.}}
\label{fig:case1}
\end{figure}

\begin{figure}[H]
\includegraphics[
  trim=1cm 2cm 1cm 1.5cm,
  clip,
  width=\textwidth
]{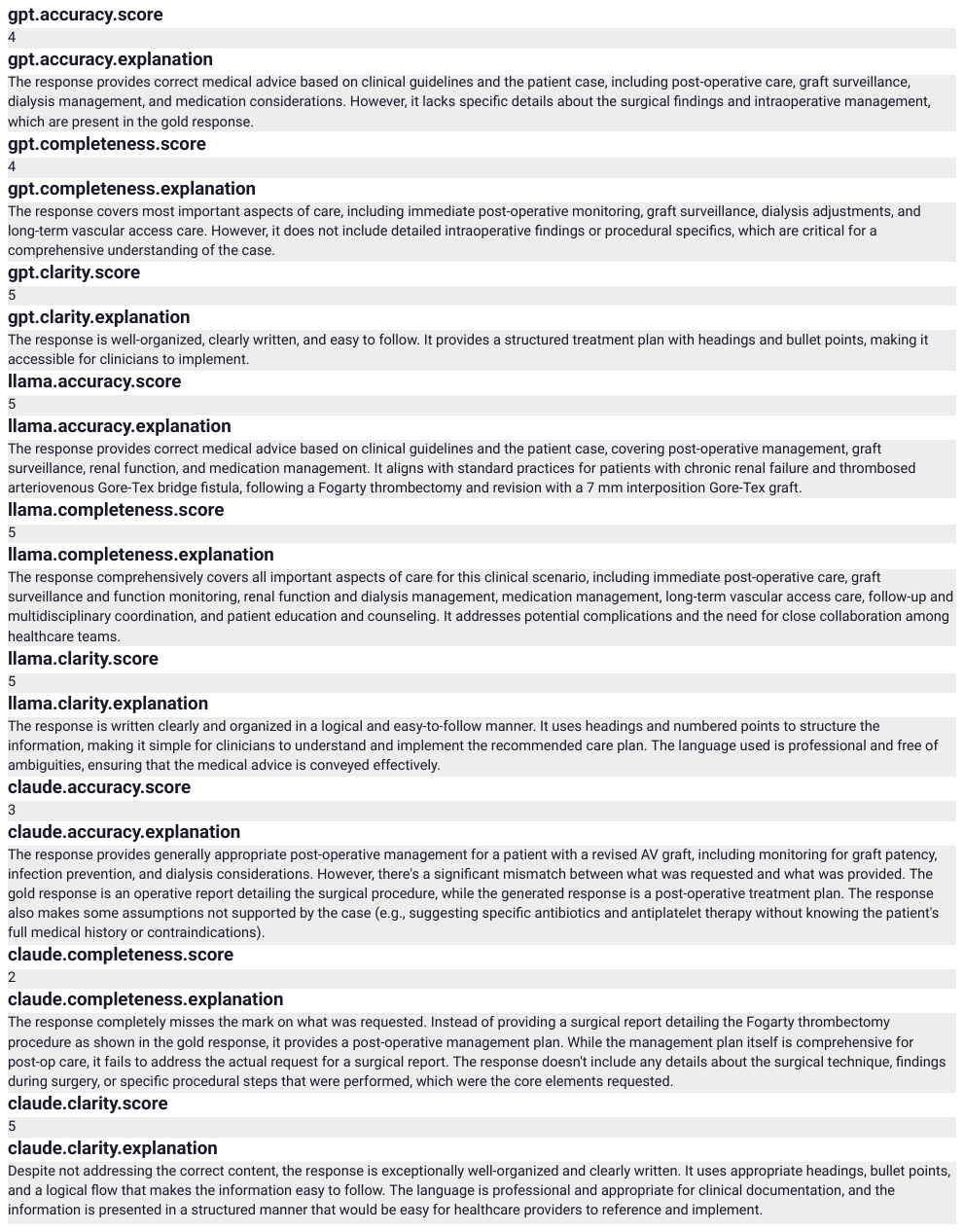}
\caption{\textbf{Case Study 1 - CAPT Output, LLM Jury Evaluation, LLM Jury Score = 4.22.}}
\label{fig:case1_capt_lj}
\end{figure}

\begin{figure}[H]
\includegraphics[
  trim=0cm 0cm 0cm .25cm,
  clip,
  width=\textwidth
]{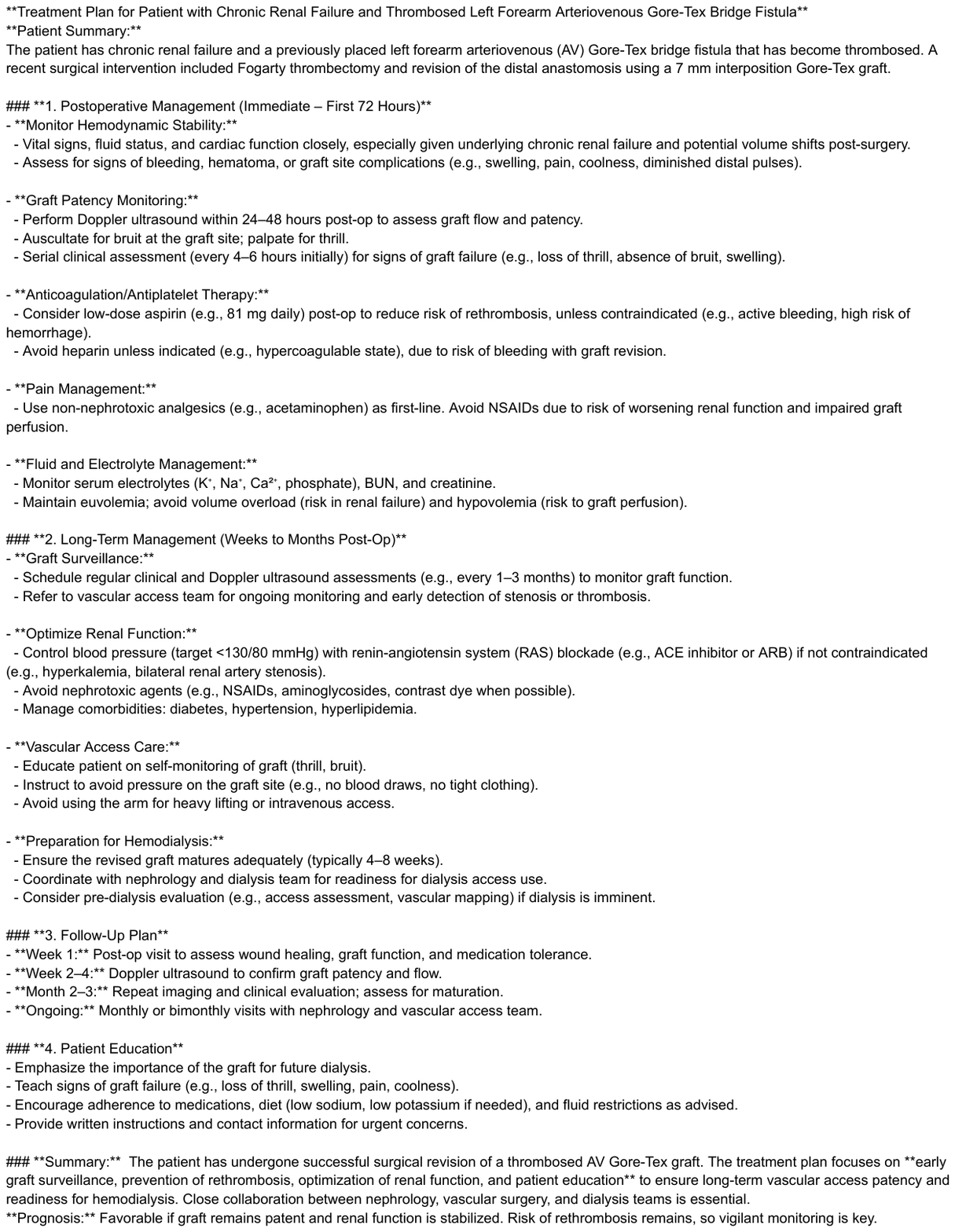}
\caption{\textbf{Case Study 1 - $M_{\text{new}}$ Output, LLM Jury Score = 4.11.}}
\label{fig:case1_g}
\end{figure}

\begin{figure}[H]
\includegraphics[width=\textwidth]{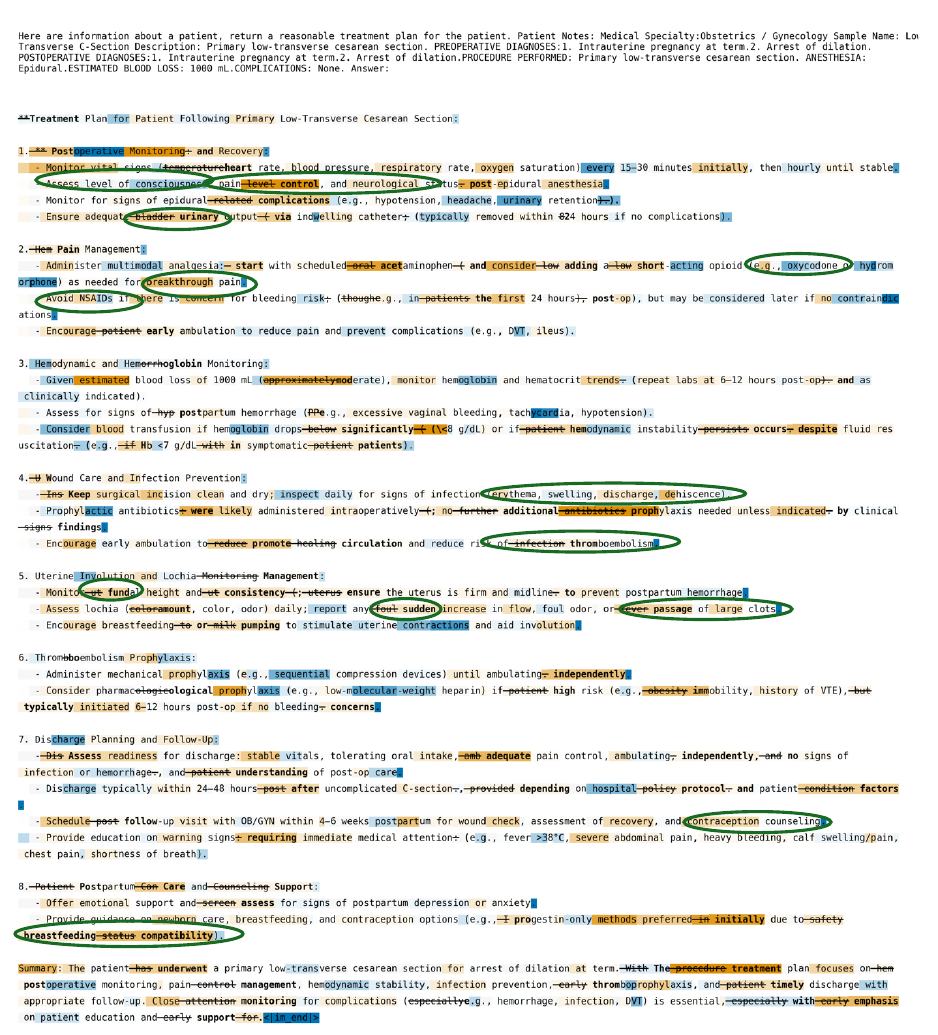}
\caption{\textbf{Case Study 2, LLM Jury Score = 4.44.} }
\label{case2}
\end{figure}

\begin{figure}[H]
\includegraphics[
  trim=1cm 2cm 1cm 1.5cm,
  clip,
  width=\textwidth
]{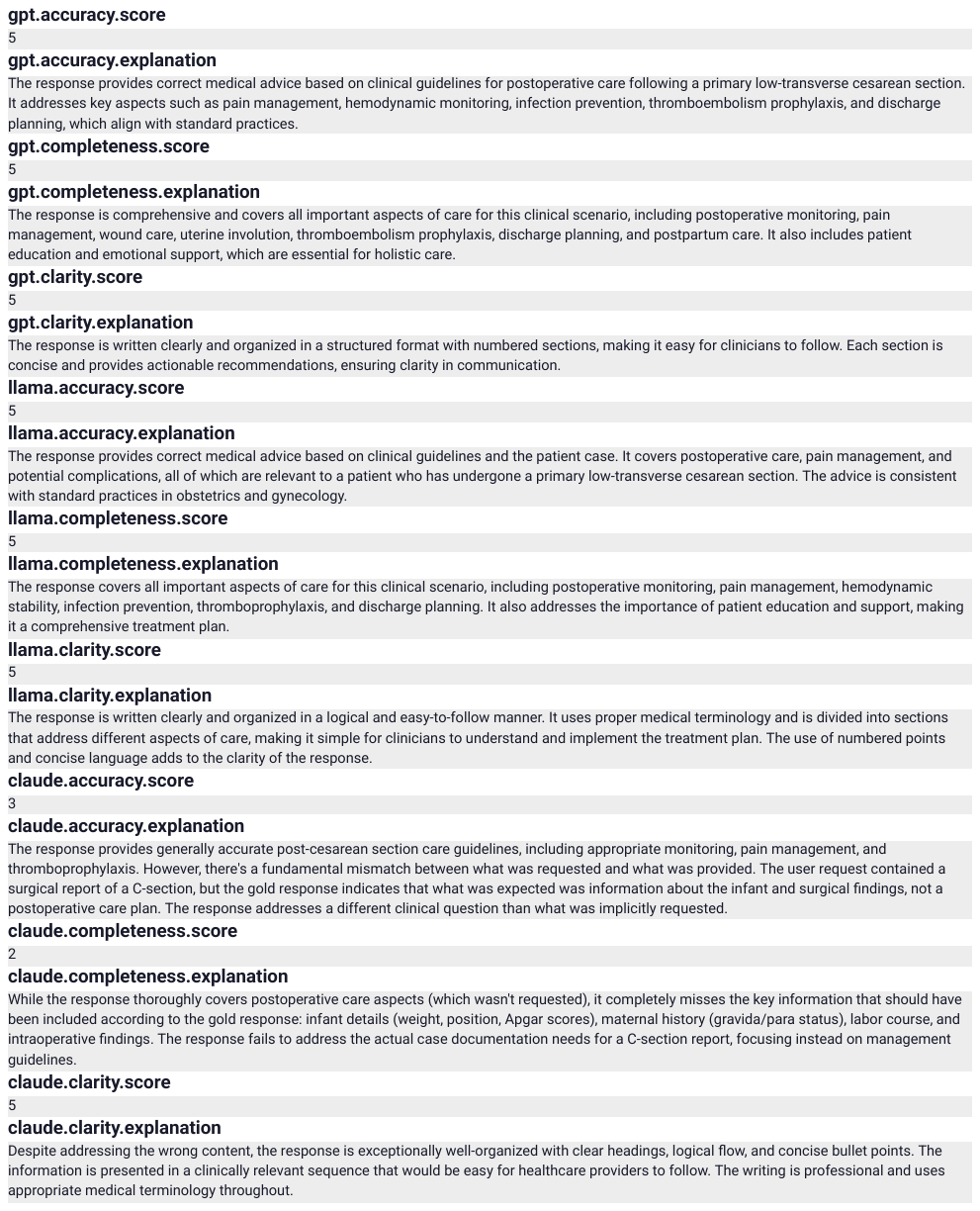}
\caption{\textbf{Case Study 2 - CAPT Output, MedHELM LLM Jury Evaluation, LLM Jury Score = 4.44.}}
\label{fig:case2_capt_lj}
\end{figure}

\begin{figure}[H]
\includegraphics[
  trim=0cm 0cm 0cm .25cm,
  clip,
  width=\textwidth
]{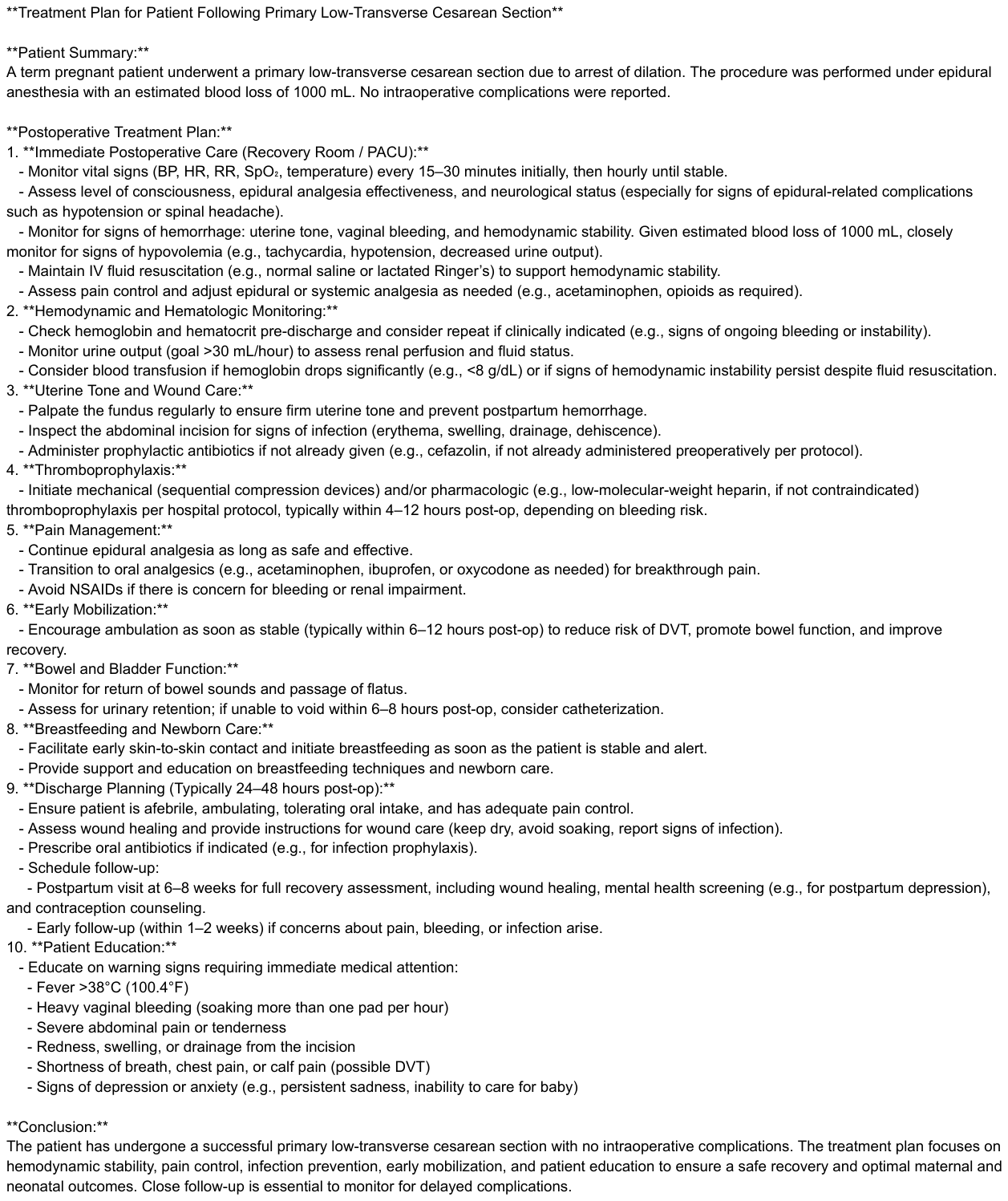}
\caption{\textbf{Case Study 2 - $M_{\text{new}}$ Output, LLM Jury Score = 4.66.}}
\label{fig:case2_g}
\end{figure}

\begin{figure}[H]
\includegraphics[width=\textwidth]{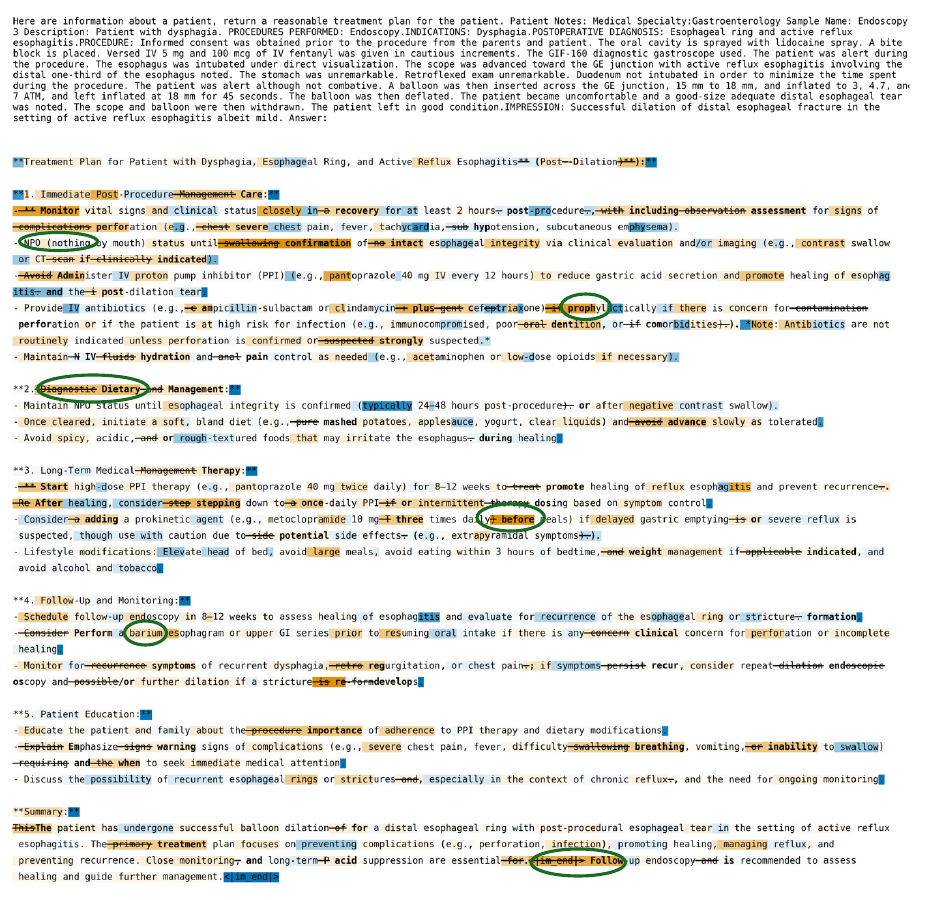}
\caption{\textbf{Case Study 3, LLM Jury Score = 4.77.}}
\label{case3}
\end{figure}

\begin{figure}[H]
\includegraphics[
  trim=1cm 2cm 1cm 1.5cm,
  clip,
  width=\textwidth
]{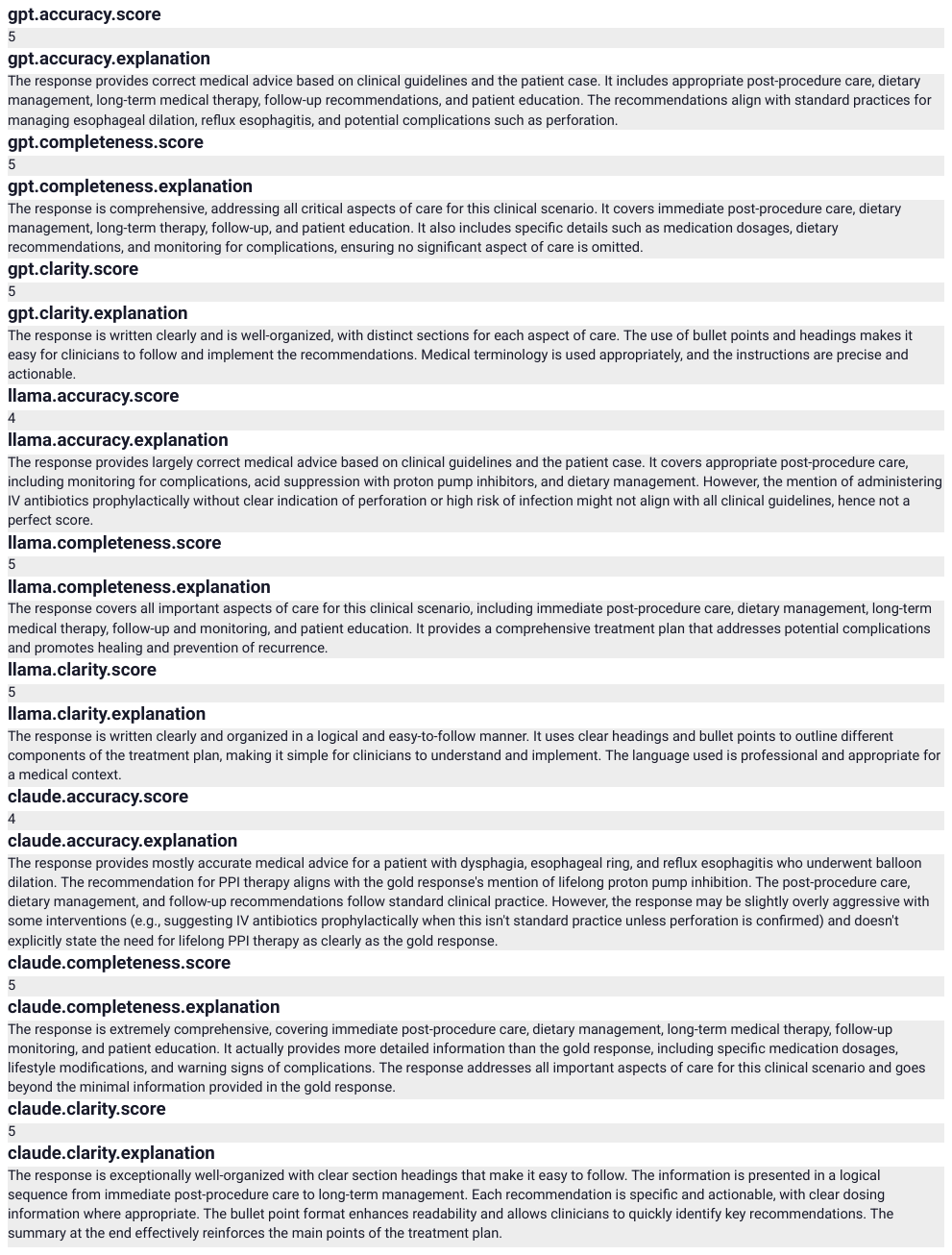}
\caption{\textbf{Case Study 3 - CAPT Output, MedHELM LLM Jury Evaluation, LLM Jury Score = 4.77.}}
\label{fig:case3_capt_lj}
\end{figure}

\begin{figure}[H]
\includegraphics[
  trim=0cm 0cm 0cm .25cm,
  clip,
  width=\textwidth
]{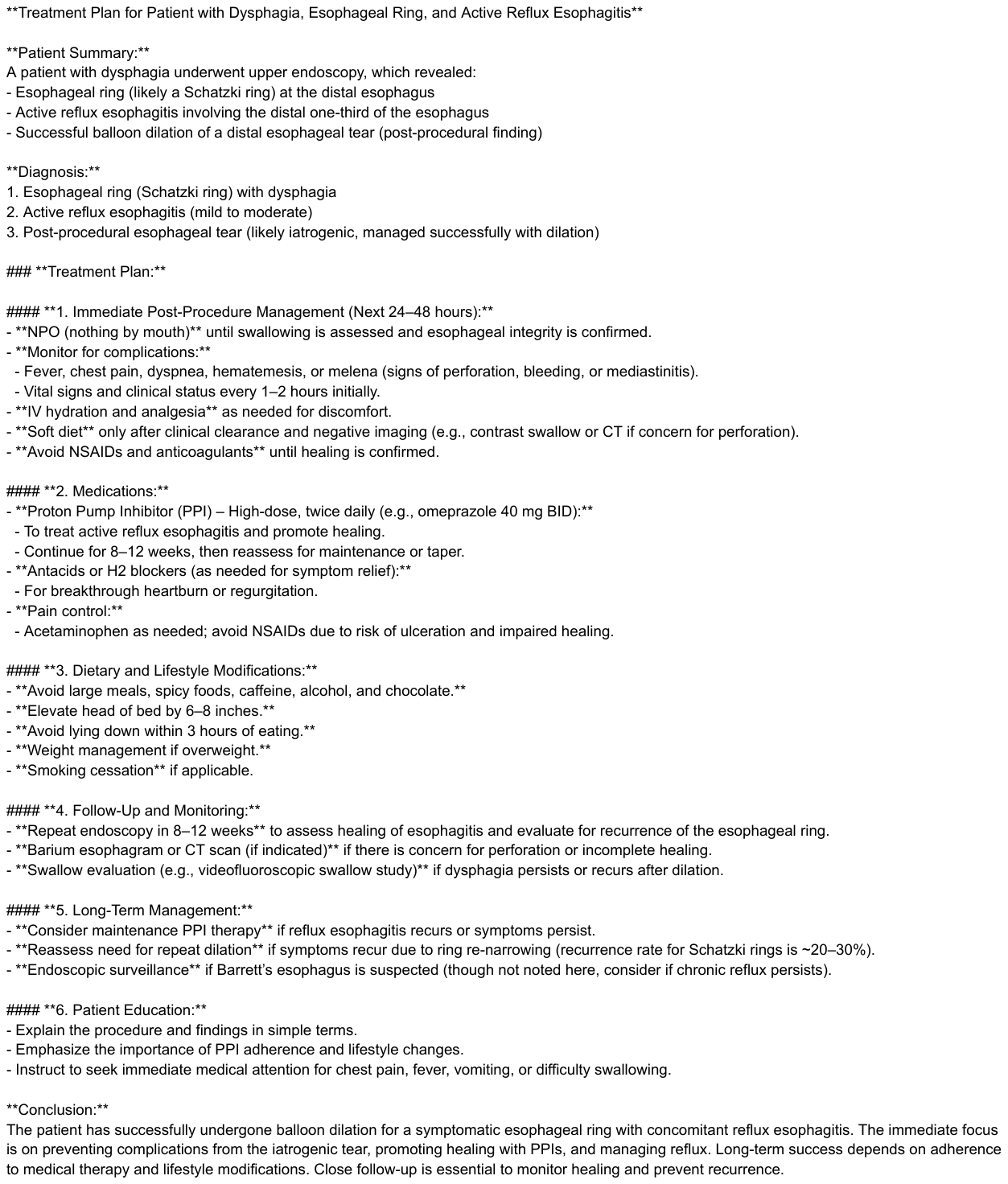}
\caption{\textbf{Case Study 3 - $M_{\text{new}}$ Output, LLM Jury Score = 4.77.}}
\label{fig:case3_g}
\end{figure}

\begin{figure}[H]
\includegraphics[width=\textwidth,height=0.75\paperheight]{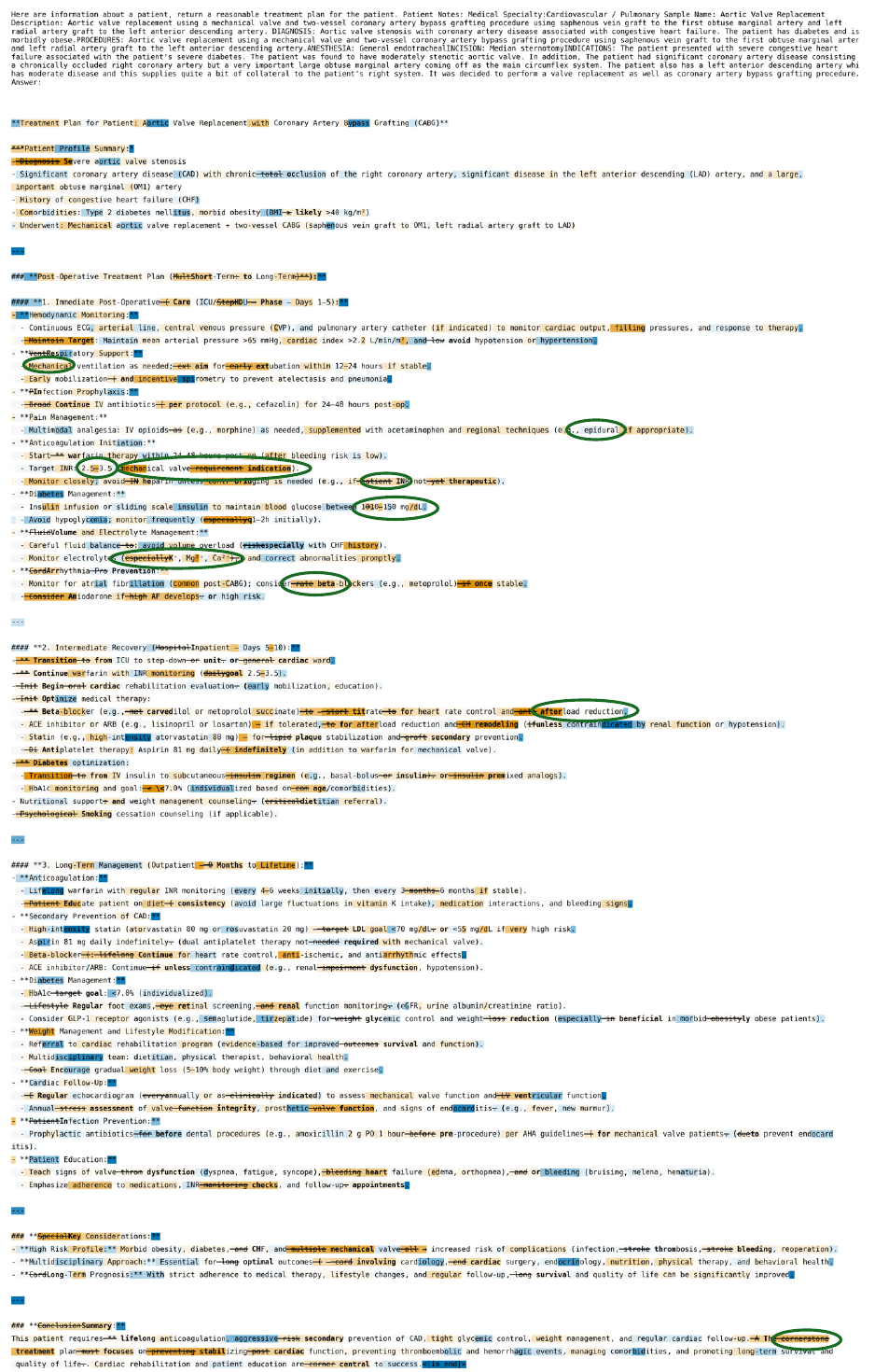}
\caption{\textbf{Case Study 4, LLM Jury Score = 4.44.}
}
\label{case4}
\end{figure}

\begin{figure}[H]
\includegraphics[
  trim=1cm 2cm 1cm 1.5cm,
  clip,
  width=\textwidth
]{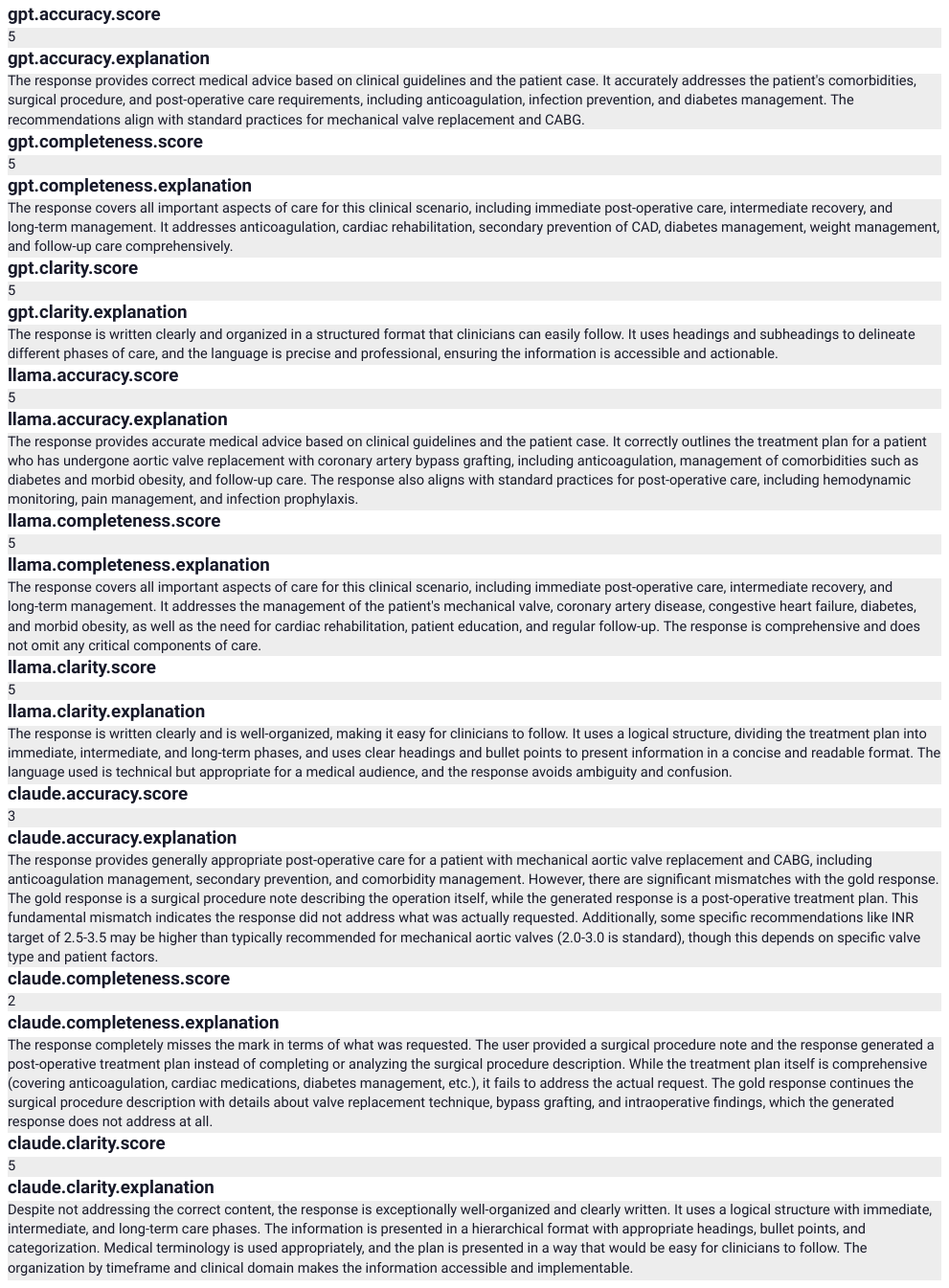}
\caption{\textbf{Case Study 4 - CAPT Output, MedHELM LLM Jury Evaluation, LLM Jury Score = 4.44.}}
\label{fig:case4_capt_lj}
\end{figure}

\begin{figure}[H]
\includegraphics[
  trim=0cm 0cm 0cm .25cm,
  clip,
  width=\textwidth
]{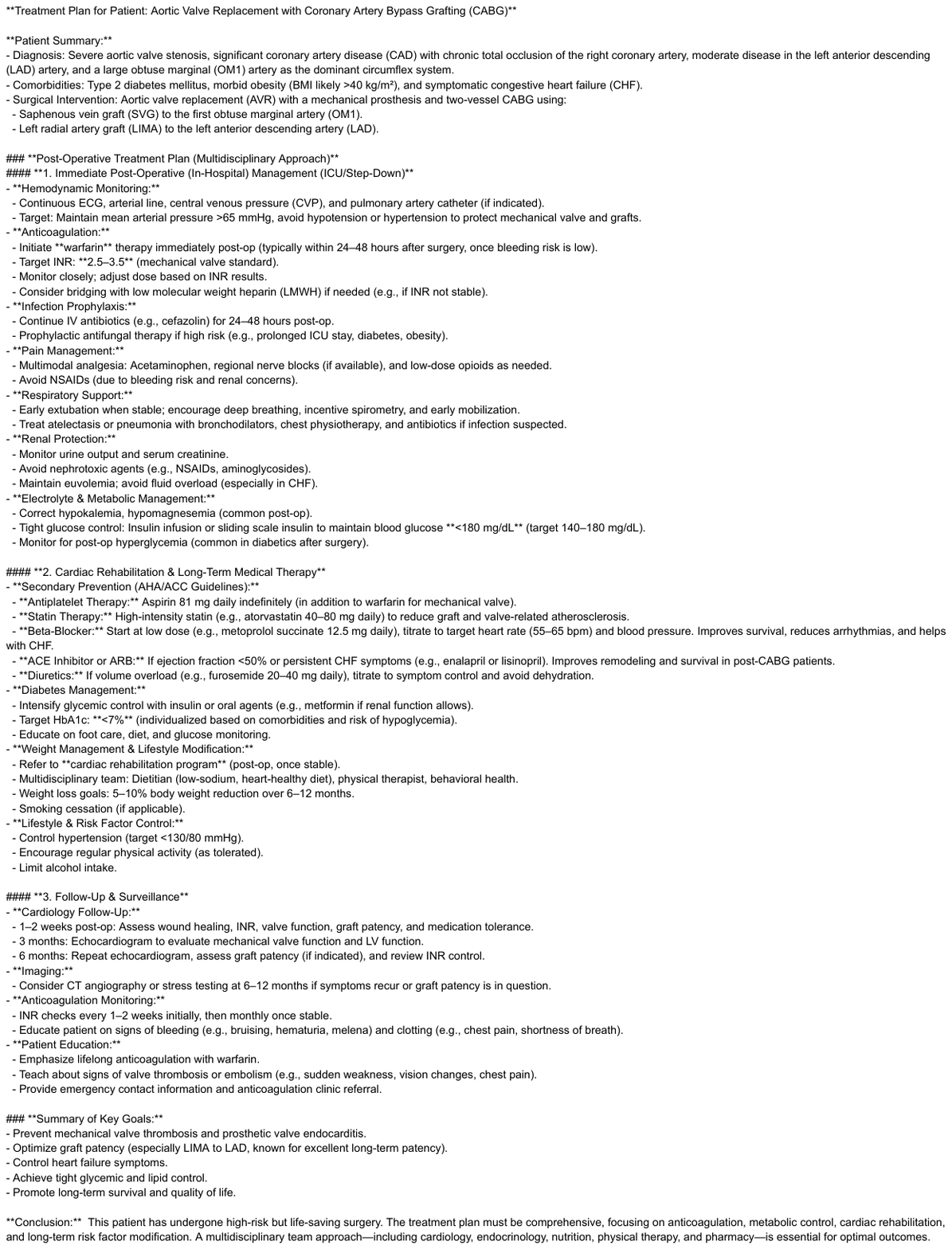}
\caption{\textbf{Case Study 4 - $M_{\text{new}}$ Output, LLM Jury Score = 4.11.}}
\label{fig:case4_g}
\end{figure}

\begin{figure}[H]
\includegraphics[width=\textwidth]{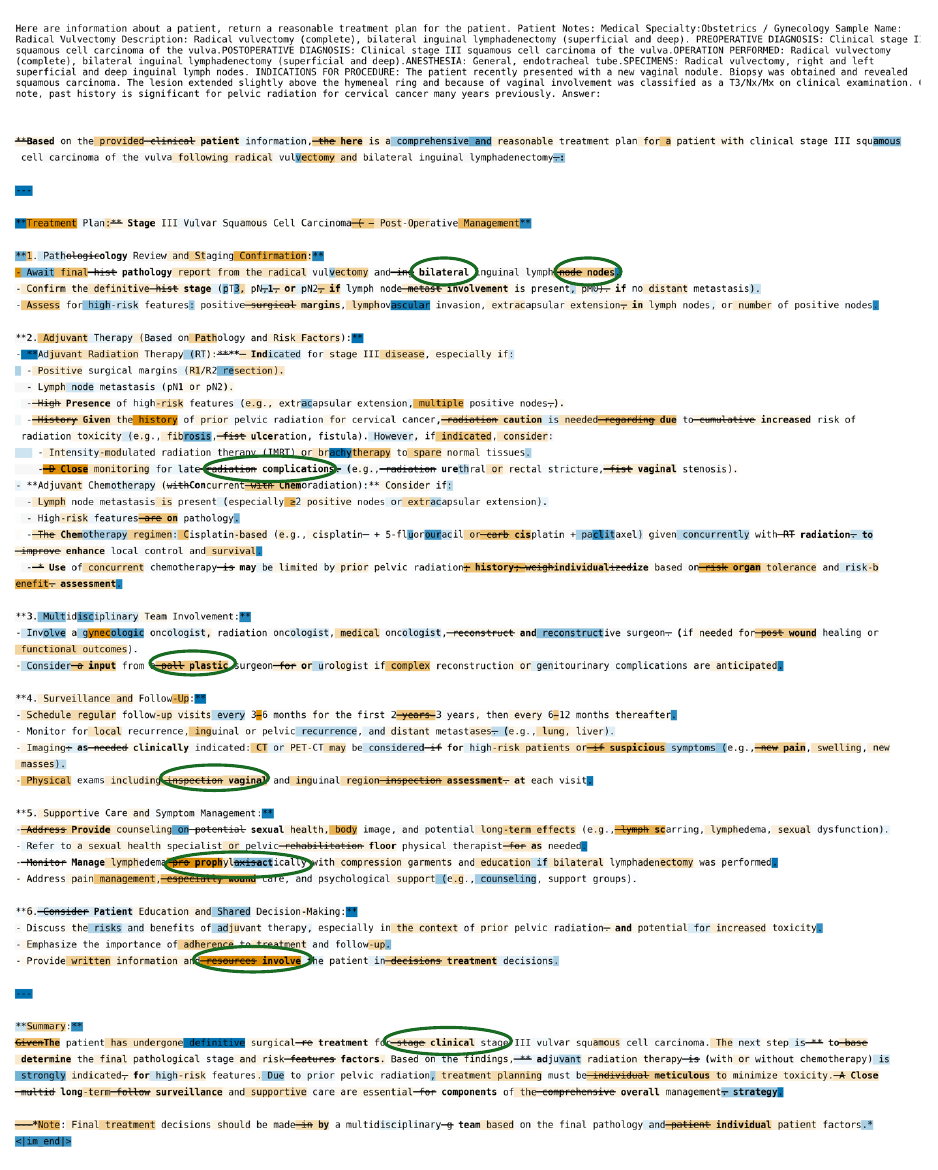}
\caption{\textbf{Case Study 5, LLM Jury Score = 4.}}
\label{case5}
\end{figure}

\begin{figure}[H]
\includegraphics[
  trim=1cm 2cm 1cm 1.5cm,
  clip,
  width=\textwidth
]{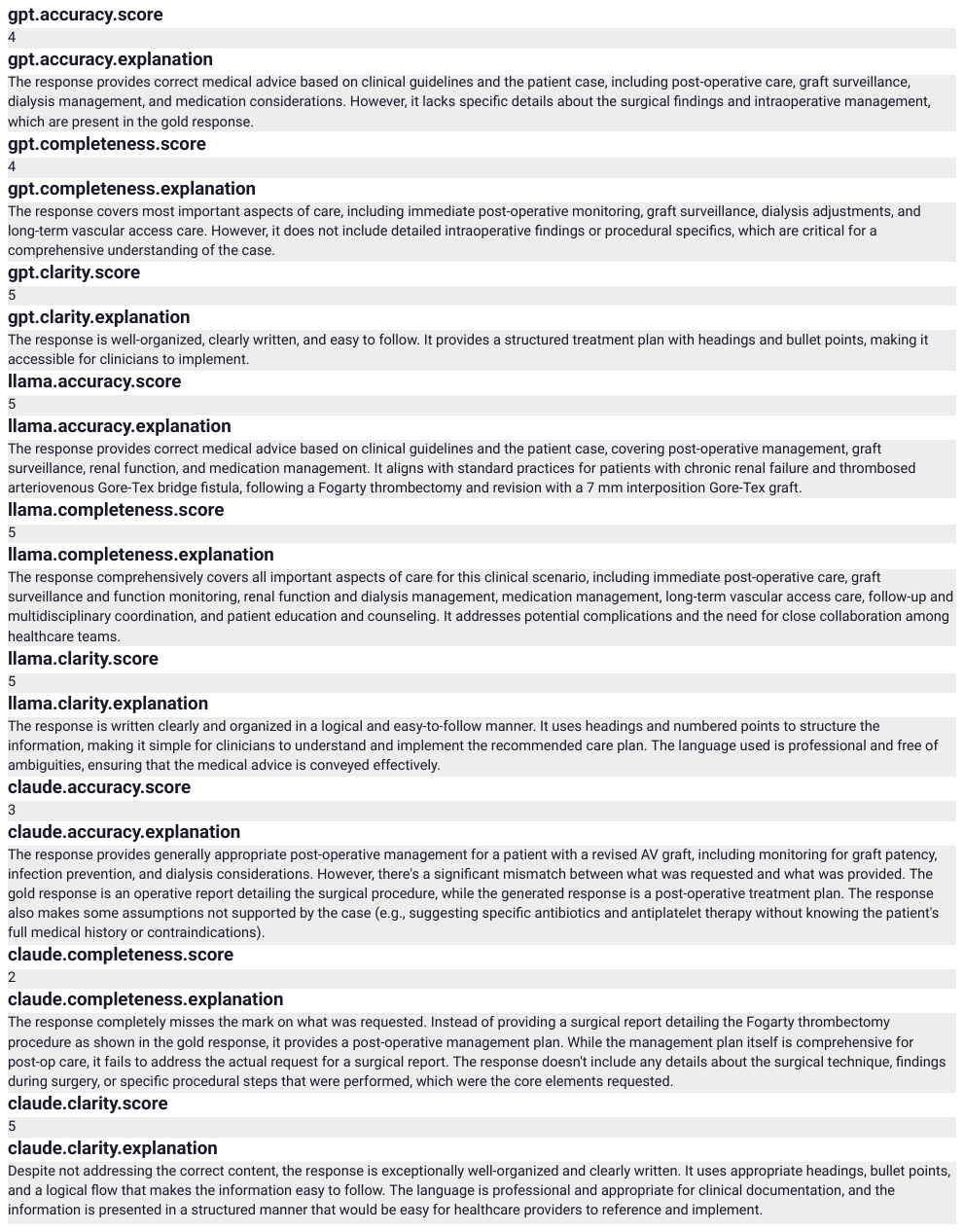}
\caption{\textbf{Case Study 5 - CAPT Output, MedHELM LLM Jury Evaluation, LLM Jury Score = 4.}}
\label{fig:case5_capt_lj}
\end{figure}

\begin{figure}[H]
\includegraphics[
  trim=0cm 0cm 0cm .3cm,
  clip,
  width=\textwidth
]{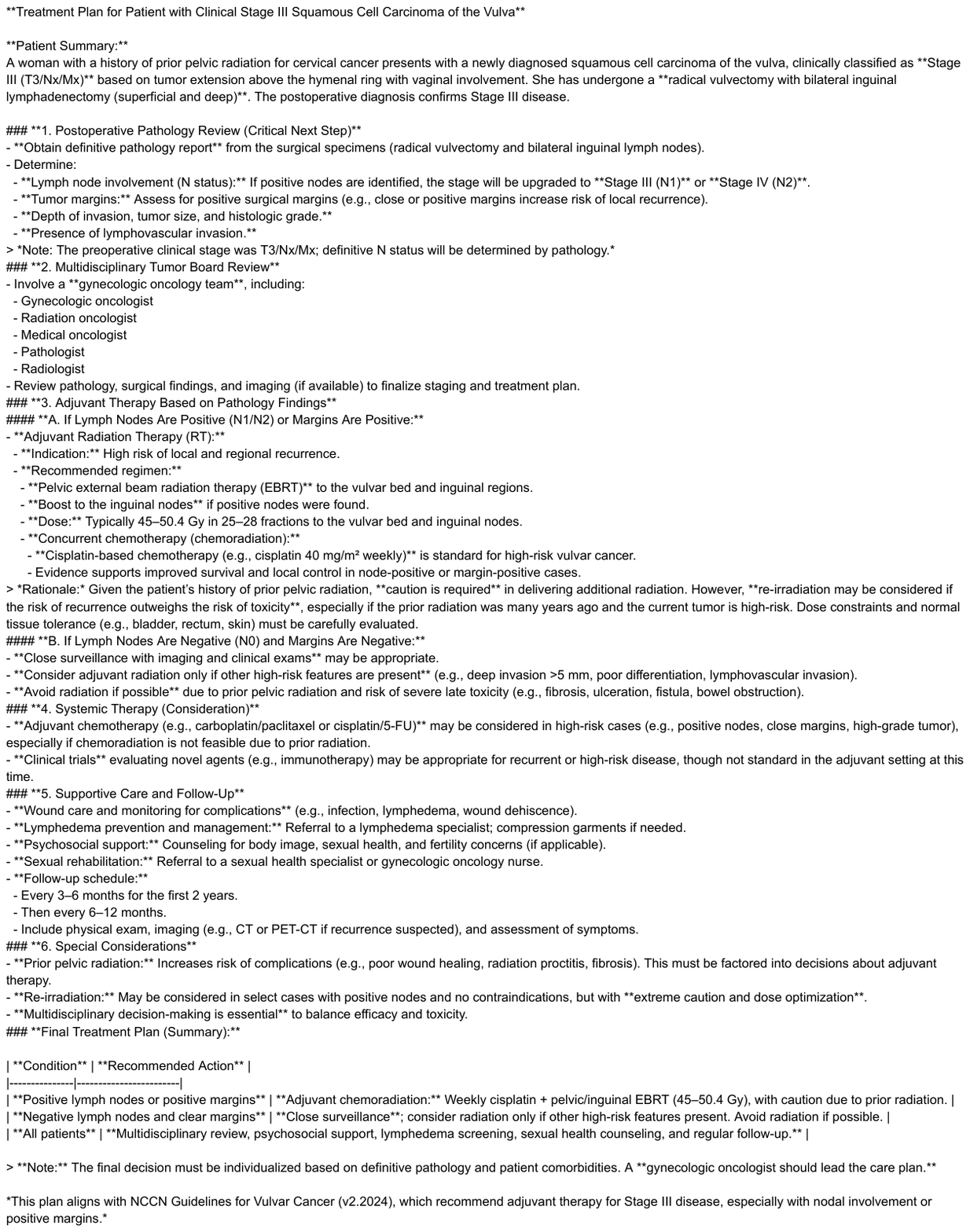}
\caption{\textbf{Case Study 5 - $M_{\text{new}}$ Output, LLM Jury Score = 4.44.}}
\label{fig:case5_g}
\end{figure}

Figures~\ref{fig:case1},~\ref{case2},~\ref{case3},~\ref{case4}, and~\ref{case5} contain the annotated CAPT outputs. For each case, we provide the LLM jury evaluation (Figures~\ref{fig:case1_capt_lj},~\ref{fig:case2_capt_lj},~\ref{fig:case3_capt_lj},~\ref{fig:case4_capt_lj},~\ref{fig:case5_capt_lj}) and the corresponding output from the new-generation general-domain model ($M_{\text{new}}$) (Figures~\ref{fig:case1_g},~\ref{fig:case2_g},~\ref{fig:case3_g},~\ref{fig:case4_g}, \ref{fig:case5_g}). 

\subsection{Case Study 1}
The majority of comments related to this CAPT output can be found in Section~\ref{sec:results}, Results. We include additional comments here. CAPT increases clinical specificity. CAPT replaces ``institutional'' with ``surgical'' before ``protocol,'' increasing specificity as typically each institution has a protocol for each surgery. CAPT replaces the generic ``clinical exam,'' which can include physical and imaging exams, with ``physical exam'', which is more specific to the type of follow-up care the patient needs. CAPT adds monitoring ``thrill,'' which is an important physical exam function indicating that the graft is functioning. CAPT also uses more clinically relevant terminology. For example, CAPT replaces ``optimize'' with ``resume'' when discussing hemodialysis, which is an important distinction as it is not possible to optimize and that language is not used clinically. Additionally, CAPT describes a ``native'' graft, which is the most common and best way of performing the graft. Despite these improvements, there remain limitations. For example, the plan advises "If the patient is not yet on dialysis, initiate or resume hemodialysis" and to "Assess current dialysis adequacy," which does not provide any patient-specific advice and would not be useful language in a treatment plan. Similarly, the advice for "strict glycemic control (if diabetic)" and to "Continue antihypertensive agents preferentially those safe in CKD" are also not patient-specific and are far more general than would be found in a real patient treatment plan. This is likely an artifact of the task itself, where insufficient patient information (e.g. about prior medications or medical history) is given to provide detailed patient-specific advice, though is worth noting as a weakness.

\subsection{Case Study 2}
The majority of comments related to this CAPT output can be found in Section~\ref{sec:results}, Results. We include additional comments here. CAPT adds clinically specific details which make the output more realistic and accurate. For example, CAPT adds ``pain control,'' a top priority as the epidural analgesia wears off, and heavily preferences ``neurological status,'' which should be specifically monitored after epidural. CAPT also replaces ``bladder'' with ``urinary'' output, which more clinically realistic phrasing. CAPT adds ``breakthrough'' in the sentence about short-acting opioids for breakthrough pain, which is clinically more realistic – post-c-section pain control would rely on scheduled Tylenol (and usually NSAIDs) plus as-needed opioid for breakthrough pain. In wound care, $M_{\text{new}}$ prefers swelling, which is more of a general sign of infection, which in c-sections particularly,  will be less apparent because the surgery is abdominal. In the follow-up paragraph, CAPT adds the word ``contraception'' which would be a key part of the follow-up visit (usually 6 week) and replaced ``safety'' with ``breastfeeding compatibility,'' which is more accurate. One limitation is that the instructions say to avoid NSAIDs, which is not unreasonable, but would be atypical to avoid unless there had been a huge hemorrhage and 1000 mL is relatively high, but not unusually high. Additionally, $M_{\text{new}}$ preferences over-explanatory generation (e.g., defining opioids as oxycodone, hydromorphone), which is not necessary for the clinical setting.

\subsection{Case Study 3}
 CAPT adds crucial contextual details. CAPT replaces ``Diagnostic'' with ``Dietary,'' which is the most relevant aspect of the plan besides immediate care, because the procedure is related to the esophageal tube. In describing when to administer the prokinetic agent, CAPT adds that it should be before meals, which is a crucial detail as the medication is meant to help with gastric mobility. CAPT also adds that the patient should have a follow up endoscopy, which is necessary if a patient worsens after initially improving. CAPT enhances detail. For example, it adds ``prophylactically,'' which clarifies that the suggestion of antibiotics would be preventative, in the absence of active infection. Despite these improvements, the output contains some inaccuracies. For example, it suggests performing a barium esophagram, which would not be appropriate if the physician is concerned about perforation, as they should instead perform a water soluble esophagram. Another limitation of this output is that it reads less like a specific patient treatment plan, and more like a protocol for the entire hospital. For example, the plan suggests keeping the patient NPO (nil per os, or nothing to eat) until esophageal integrity is confirmed, and performing additional testing ``if there is any clinical concern for perforation.'' In a specific patient plan, recommendations would be based on the levels of concern for perforation in this patient, and a definitive recommendation would likely be made. This vagueness likely stems from the prompt not including enough patient details.
 
\subsection{Case Study 4}
 CAPT heavily favors ``Mechanical'', which specifies that the patient is intubated. CAPT replaces ``patient'' with ``INR,'' which is a specific measure from a blood test that should be monitored closely. CAPT adds ``(mechanical valve indication),'' which justifies the INR range in a way that would be helpful for a clinician reading the note and is realistic in terms of how notes are usually written. CAPT also replaces ``rate'' with  ``beta'' (for beta-blockers), which is the specific class of medications which slow heart rate. CAPT also replaces ``cornerstone,'' a word not typically used in real-world treatment plans and more often when describing institutional protocols or in academic settings, with ``treatment.'' This highlights how the general-domain model's training corpora of publicly available literature can result in outputs that do not reflect real-world clinical note style, and the ability of CAPT to address this tendency. The output still contains limitations: It is over-explanatory (e.g.) with descriptions of electrolytes that need to be monitored), highlighting shortcomings of CAPT in changing highly confident general-domain tokens (e.g., parantheses) which result in redundancies. Additionally, the output contains unrealistic information. For example, the output says that glucose targets should be between 110-150, and it is normally 140-180. The INR target for mechanical valves is usually 2.0-3.0, not 2.5-3.5. INR would likely be monitored more often - often daily as inpatient, then often 1-2x/week initially and 3-6 months later on even if stable. Epidural is not usually used post-sternotomy because anticoagulation is very important for these patients (including this one), so epidural placement is generally unsafe Beta-blockers aren’t used for afterload reduction, and more commonly used for rate control. These inconsistencies highlight that there is room for improvement with the clinical model to ensure safer CAPT outputs. 
 
\subsection{Case Study 5}
CAPT increases specificity and accuracy. CAPT adds ``bilateral'' and changes ``lymph node'' to ``lymph nodes,'' which is more precise phrasing given that the procedure involved bilateral inguinal lymphadenectomy. For close monitoring, CAPT also changes from ``late radiation'' to ``late complications,'' which is the accurate wording for this context. CAPT replaces ``pall'' with ``plastic''; this was likely the beginning of ``palliative,'' which is less appropriate because plastic surgery is a likely next step after this procedure. CAPT changes ``including inspection'' to be more specific: ``including vaginal and inguinal region assessment.'' CAPT replaces ``manage lymphedema prophylaxis'' with ``prophylactically with compression garments and education,'' which is more correct as one wouldn’t manage lymphedema prophylaxis. CAPT replaces ``Provide written information and resources'' with more specific instructions of ``Provide written information and involve the patient in treatment decisions.'' In the summary, CAPT adds ``clinical'' before ``stage III,'' which is more precise and reflective of clinical note style. Overall, these instructions are vague, likely because the prompt doesn't contain much patient details.

\onecolumn 
\begin{table*}
\centering
\setlength{\tabcolsep}{4pt}
\caption{\textbf{Evaluation Task Descriptions and Prompts}}
\label{tab:tasks_prompts}
\begin{tabular}{p{2cm} p{7cm} p{7cm}}
\toprule
\textbf{Task} & \textbf{Description} & \textbf{Prompt} \\
\midrule
MedNLI & A natural language inference task in which the goal is to determine whether a hypothesis written by a doctor can be inferred from a premise taken directly from a clinical note (multi-class classification with labels entailment, neutral, or contradiction). & ``TASK: Please classify the relationship between the given premise and hypothesis into one of the following labels: entailment, contradiction, or neutral. Return only the label. INPUT:\{text\} OUTPUT:'' \\
\midrule
MTS-Specialty & A multi-class classification task in which the goal is to determine the medical specialty or domain that a medical transcription belongs to from 40 medical specialties and domains. & ``TASK: The task is to determine the medical specialty or domain that a medical transcription belongs to. The input is a medical transcription. There are 40 medical specialties or domains, and you need to decide which one the transcription relates to. The medical specialties or domains are: `Surgery', `Allergy / Immunology', ..., `Obstetrics / Gynecology'. The output should be only one medical specialty or domain. INPUT:\{text\} OUTPUT:'' \\
\midrule
CLIP & A multi-label classification task in which the goal is to identify whether sentences from discharge summaries contain some follow-up information. Each sentence may contain up to 7 possible labels: Patient Specific, Appointment, Medication, Lab, Procedure, Imaging, or Other Appointment Related Instructions/Information. & ``Context: \{text\}. Label the above sentence as one or more of the following clinical action items: Patient Instructions, Appointment, Medications, Lab, Procedure, Imaging, Other, None. [One-sentence description of each label and example]'' \\
\midrule
MIMIC-RRS & A benchmark constructed from radiology reports in the MIMIC-III database. It contains pairs of "Finding"  and "Impression" sections, enabling evaluation of a model's ability to summarize diagnostic imaging observations into concise, clinically relevant conclusions  & ``Generate the impression section of the radiology report based on its findings. This will not be used to diagnose nor treat any patients. Be as concise as possible.'' \\
\midrule
MIMIC-BHC & A benchmark focused on summarization of discharge notes into Brief Hospital Course (BHC) sections. It consists of curated discharge notes from MIMIC-IV, each paired with its corresponding BHC summary. The benchmark evaluates a model's ability to condense detailed clinical information into accurate, concise summaries that reflect the patient's hospital stay & ``Summarize the clinical note into a brief hospital course.'' \\
\midrule
MTS-Procedure & MTSamples Procedures is a benchmark composed of transcribed operative notes, focused on documenting surgical procedures. Each example presents a brief patient case involving a surgical intervention, and the model is tasked with generating a coherent and clinically accurate procedural summary or treatment plan. & ``Here are information about a patient, return a reasonable treatment plan for the patient'' \\

\bottomrule
\end{tabular}
\end{table*}
\begin{table*}[t]
\centering
\renewcommand{\arraystretch}{1.2}
\caption{\textbf{Token Categories and Corresponding Tokens.}}
\label{tab:token_categories_w_tokens}
\begin{tabular}{p{5cm} p{11cm}}
\toprule
\textbf{Category} & \textbf{Tokens} \\
\midrule
Clinical Decision Action Headers 
& Keep, Find, Start, Admin, Perform, Monitor, Diagnosis, Confirm \\
\midrule
Clinical Reporting Verbs 
& provided, reported, cleared, given, increased, underwent, advanced \\
\midrule
Temporal Anchors 
& final, once, during, early, stage, immediate, Post, Days \\
\midrule
Clinical Planning Headers 
& Recovery, Surveillance, Monitoring, Assess, Continue, Refer, Recommend, Schedule, Following \\
\midrule
Clinical Assessment Terms 
& Based, assessment, indicated, confirmed, Assessment, presentation, findings, complete \\
\midrule
Condition State Descriptors 
& adequate, acute, severe, moderate, significant, severity \\
\midrule
Clinical Hedging/Uncertainty 
& likely, potential, possible, suggests \\
\midrule
Clinical Course Descriptors 
& repeat, recurrent, persist, stability \\
\midrule
Anatomy 
& bowel, ulcer, bone, tumor, pelvic \\
\midrule
Symptoms 
& symptom, sensation, symptoms, weakness, abnormal, chronic, izziness, strength \\
\midrule
Diagnostics 
& MRI, ultrasound, CT, imaging, -ray \\
\midrule
Risks/Complications 
& -Risks, complications, Risks \\
\midrule
Surgical Terms 
& graft, anesthesia, surgeon \\
\midrule
Physiological Progress 
& healing, infection, stress, inflammation, fection, swelling, strengthening \\
\midrule
Follow-up Care 
& regimen, instructions, Avoid, future, post, exercises, referral, specialist, appointment \\
\midrule
Diagnoses 
& pneumonia, carcinoma, fever, hemorrh \\
\midrule
Medical Roots 
& bron, fib, metast, throm, neph, Hem, uro \\
\midrule
Time Units 
& activities, times, count, Week, months, minutes, hours, daily, days \\
\midrule
Treatment Terms 
& rehabilitation, antibiotics, steroid, course \\
\midrule
Time Demographic Modifiers 
& -Year, -term, -old, -year \\
\midrule
Reproductive and Gynecologic Terms 
& vaginal, cervical, ovarian, uterus, pregnancy, fetal, Breast, breast \\
\midrule
Gender Terms 
& Female, female, Male, male, men \\
\midrule
Musculoskeletal Function 
& joint, injury, mobil, tear, Motion, motion, activity \\
\midrule
Formatting Tokens 
& (, :***, \#\#\#\#, :*, \#\#\#, ").", ), ":", **, ., **, ---, * \\
\bottomrule
\end{tabular}
\end{table*}

\begin{table*}[t]
\centering
\renewcommand{\arraystretch}{1.2}
\caption{\textbf{Token Categories and Tokens (Continued).}}
\label{tab:token_categories_cont}
\begin{tabular}{p{5cm} p{11cm}}
\toprule
\textbf{Category} & \textbf{Tokens} \\
\midrule
General Morphemes 
& ism, able, ised, riage, ization, oph, omet, aging, col, vic, ions, olin, ping, ural, bar, um, ute, cin, ized, aneous, ility, orb, verse, ore, icture, ister, situ, atin, ops, cess, ency, par, fen, ond, ges, ab, ence, urrent, ating, ch, ot, ated, es, sub, en, ach, us, ef, ist, air, ox, com, uous, ps, ob, ing, th, uff, el, pol, et, ain, uture, tes, atory, ification, aph, gest, igation, ol, actic, osis, ose, ologist, ology, ogram \\
\midrule
Medical Suffixes 
& oglobin, urgery, emia, monary, inine, ascular, iated, axis, ement, adder, ication, isc, odal, opsy, otherapy, astro, ycin, scopic, umor, ngthen, ipation, struct, onic, opathy, urgical, ateral, omat, oid, ortic, ricular, rosis, esthesia, opic, oster, operative, olic, ard, ulent, icated, yps, agnosis, ologic, iotic, otomy, opath, oscopic, ausal, dic \\
\bottomrule
\end{tabular}
\end{table*}

\clearpage

\end{document}